\documentclass{article}

\usepackage[preprint]{neurips_2026} 

\usepackage[utf8]{inputenc} 
\usepackage[T1]{fontenc}    
\usepackage{hyperref}       
\usepackage{url}            
\usepackage{booktabs}       
\usepackage{amsfonts}       
\usepackage{nicefrac}       
\usepackage{microtype}      
\usepackage{xcolor}         

\usepackage{lipsum}
\usepackage{amsmath,amsthm}
\usepackage{algorithmic}
\usepackage{algorithm}
\usepackage{array}
\usepackage{textcomp}
\usepackage{stfloats}
\usepackage{verbatim}
\usepackage{graphicx}
\usepackage{mwe}
\usepackage{subcaption}
\usepackage{float}
\usepackage{multirow}
\usepackage{lscape}
\usepackage{listings}

\title{GenAutoML: An Agentic Framework for Dynamic Architecture Generation and Optimization in Time-Series Analysis}

\author{
Oleeviya Babu Poikarayil$^{1}$, Cédric Schockaert$^{1}$, Abdulrahman Nahhas$^{2}$,\\
Christian Daase$^{2}$, Mursal Dawodi$^{3}$, \textbf{Jawid Ahmad Baktash}$^{3}$\\[0.5ex]
$^{1}$Paul Wurth S.A., Luxembourg\\
$^{2}$Otto-von-Guericke University, Magdeburg, Germany\\
$^{3}$Technical University of Munich, Germany\\[0.5ex]
\texttt{oleevyazehr@gmail.com, cedric.schockaert@sms-group.com}\\
\texttt{\{abdulrahman.nahhas, christian.daase\}@ovgu.de}\\
\texttt{\{mursal.dawodi, jawid.baktash\}@tum.de}
}
\begin{document}
\maketitle

\begin{abstract}
Designing neural architectures for time series forecasting and anomaly detection is a resource-intensive task requiring extensive domain knowledge. Traditional Automated Machine Learning (AutoML) systems typically rely on static, pre-defined search spaces, which fail to adequately account for the variability of datasets. We present GenAutoML, an agentic framework that utilizes Large Language Models (LLMs) to act as neural architects in order to bridge the gap between natural language requirements and executable PyTorch code. The system features a closed-loop architecture that includes a Sandboxed Reflection Loop for autonomous code repair as well as a Signature Aware Runtime that enforces architectural compliance. Additionally, we introduce a Dynamic Reversible Instance Normalization (Dyn-RevIN) wrapper to enhance robustness against non-stationary data. Experimental testing across ETTh1, ETTm1, and Weather datasets demonstrates that the framework dynamically generates novel topologies tailored to the data. While massive foundation models achieve high accuracy at the cost of severe latency, the agent-generated WaveInterferenceNet achieves a state-of-the-art inference latency of <0.01ms per sample.
\end{abstract}

\section{Introduction}

The Stationarity Problem (also known as the Stationarity Paradox), has continued to be a barrier in industrial environments for multi-variate time-series data analysis (e.g. monitoring the power grid, industrial IoT) \cite{liu2022non}. While numerous deep learning techniques such as LSTM networks, Transformers, etc. \cite{lstm, vaswani2017attention} have produced strong performance results but they perform poorly when applied to real-world datasets like the Electricity Transformer Temperature datasets (ETTh1, ETTm1) \cite{zhou2021informer} and the Weather benchmark due to their non-stationary charectristics and/or complex temporal interdependencies. The traditional model-building process follows a Static Design Method where engineers select their architecture and manually adjust hyperparameters of their designs. This manual work can takes significant amount of time to complete and could lead to the production of models that were built upon inadequate inductive bias.

Both Automated Machine Learning (AutoML) \cite{automl} and Neural Architecture Search (NAS) \cite{elsken2019neural} have been developed to eliminate manual bottlenecks in the creation of machine learning models. 
The current frameworks of NAS are constrained to predefined architectural spaces (i.e., a fixed forms of interconnections and predetermined number of layers) and do not have any kind of ``Semantic Creativity'' to invent new structural components (e.g., designing a custom temporal mixing layer instead of relying on a standard Residual connection). As a result, there is often a very substantial "design-to-deployment" gap when attempting to put an entirely new model into production, since it typically requires a complete system restart.

To address these limitations, we present GenAutoML, an agentic framework that changes the paradigm from "model search" to "Agentic Neural Engineering." Our framework considers the AI not merely as a model that can be used for optimization, but as an autonomous "Neural Architect" that is capable of reasoning about data requirements and constructing the inherent structural solutions \cite{wang2024survey}, \cite{shinn2023reflexion}.

\paragraph{Differentiation from Time-Series Foundation Models} 
Concurrently, the field has seen a rapid paradigm shift toward zero-shot Time-Series Foundation Models (TSFMs) such as Chronos \cite{ansari2024chronos} and MOIRAI \cite{woo2024unified}. However, these gigantic foundation models (often exceeding 100M+ parameters) require enormous amounts of computational resources and result in high inference latency. For this reason, they cannot be used in real-time, resource limited environments like Edge AI (for example, embedded industrial sensors, power grid monitoring) \cite{lin2020mcunet}, \cite{merenda}. In order to address this limitation, GenAutoML is explicitly designed for the Edge. We shift the paradigm from "massive model search" to "Agentic Neural Engineering," focusing on \textbf{deterministic edge-deployable synthesis}. Rather than solely chasing state-of-the-art predictive accuracy at the cost of massive parameter counts, our framework dynamically invents ultra-lightweight architectures (e.g., under 1 million parameters). These generated topologies trade marginal fractions of accuracy for microsecond inference latency (<0.01 ms) and highly stable deployment characteristics.

In summary, the main contributions of our work are as follows:

\textbf{Agentic Neural Synthesis with Sandboxed Reflection (The Semantic Interface):} We present an LLM based reasoning engine that translates natural language domain constraints into executable PyTorch structural definitions. GenAutoML creates a search environment for the discovery of inductive biases that are unique to the specific problem domain through direct mapping of semantic needs (e.g., capture the temporal patterns of location over multiple levels of granularity') to architectural elements. In addition, to ensure that these structures have a mathematical form of validity, we proposed a Sandboxed Reflection Loop, that executes the generated code using simulated tensor structures within an in-process validation. By feeding back the traceback and dimension mismatch errors to the LLM used to generate them, the Sandboxed Reflection Loop iterates through the process to help debug the architecture in preparation for final deployment \cite{chen2023teaching}.

\textbf{Runtime Architectural Injection:} We were able to create a reflection-driven orchestration pipeline that allows for the real time injection of generated neural modules into an active optimization runtime environment \cite{paszke2019pytorch}. The ability to seamlessly perform "Hot-Swaps" in real time allows us to validate and enhance newly designed architectures without disrupting any of the training or search processes.  In addition, using Signature Aware Runtimes and Shape-Agnostic Projection Heads prevents catastrophic failures from occurring when executing LLM code signatures that weren’t defined previously. This means that the system will be able to dynamically adjust to any defined forward function and automatically project all unmatched latent channels to their corresponding target dimensions.

\textbf{Mathematical Hardening via Dynamic RevIN (The Safety Guardrail):} We have integrated a Dynamic Reversible Instance Normalization (Dyn-RevIN) framework in order to bridge the gap between autonomous code generation and production stability, \cite{kim2022reversible}. The Dyn-RevIN Framework provides a common statistical envelope for separating operational/architectural logic from output’s numerical stability requirements that the agent will develop. Furthermore, the Dyn-RevIN Framework imposes stationary constraints on the structure of the input and output (IO) providing the convergence of results among all architectures including unverified architectures for highly non-stationary signals. The Dyn-RevIN Framework maintains continuous evaluation integrity throughout the entire evaluation cycle by prohibiting access to temporal data throughout the entire evaluation time frame using asymmetric data regularization.

\textbf{End-to-End Conversational Pipeline:} GenAutoML provides a conversational interface for performing machine learning workflows. The workflow proceeds as follows: (1) The user first uploads multivariate datasets (CSV and Parquet) into the conversational interface. (2) A LangChain-powered Pandas DataFrame agent functions as a virtual data scientist to answer exploratory queries via Python code that is executed at the user’s command, allowing for interactive visualizations to be returned. (3) Then the user can tell the agent to design a custom neural architecture (e.g., "Design a Lightweight Inception model"). (4) Prior to Injection of the architecture into a live, Optuna-based pipeline via a dynamic runtime loader for hyperparameter optimization and training, the user verifies the correctness of the generated PyTorch code by executing it in a in-process validation harness version of the original code \cite{akiba2019optuna}.

\section{Related Work}
\label{sec:related_work}

Using the PRISMA framework, we conducted a Systematic Literature Review (SLR) to ensure a comprehensive and reproducible evaluation of the current  state of research. We performed a boolean search in the Scopus database to capture the rapidly advancing intersection of generative AI and architecture search using the following query limited to publications after 2022: \texttt{TITLE-ABS-KEY ( ( "Neural Architecture Search" OR "NAS" OR "AutoML" ) AND ( "Large Language Model" OR "LLM" ) AND ( "code generation" OR "topology" OR "PyTorch" OR "code" ) )}. 

 Through our preliminary search, we retrieved a total of 59 records. Following preliminary filtering, 34 records were selected based on the title and abstract review. From these, 30 documents remained eligible based on their clear focus upon the use of large language model powered machine learning pipelines, multi-agent topology design, and the use of reflective feedback loops within the design and implementation of generative machine learning-based systems. To achieve the goal of maintaining a cohesive story within a strict conference page constraints, we distilled these 30 papers to the 11 most directly relevant studies to be included in our final synthesis. These core papers, categorized below, form the precise theoretical foundation and comparative landscape for the GenAutoML framework.

\subsection{Automated Machine Learning and Neural Architecture Search}
Traditional Automated Machine Learning (AutoML) \cite{hutter2019automated} and Neural Architecture Search (NAS) \cite{elsken2019neural} have made significant improvements in the manual barriers of developing machine learning models. Recently, there has been a shift in the field with a movement towards using Large Language Models (LLMs) to fully automate the end-to-end pipelines \cite{llm_end_to_end, llm2automl, automl_agent, uniautoml}. For example, Works such as LLMatic \cite{llmatic} and LLM-Guided Evolution \cite{llm_guided_evo} demonstrate that LLMs can act as evolutionary optimizers, which introduce meaningful variations to neural network code rather than navigating static, predefined search spaces. In addition, proposals such as NADER \cite{nader}, incorporated multiple agents that have been proposed to iteratively design and enhance base architectures. However, these approches typically output static code blocks or graph representations that require the rebooting of the complete pipeline to evaluate. GenAutoML introduces a Just-In-Time (JIT) dynamic loader that injects PyTorch \texttt{nn.Module} classes directly into a live Optuna instance, enabling seamless integration using hot-swapping without restarting the active search environment.

\subsection{Large Language Models for Autonomous Code Generation}
Recent works has shown that LLMs are having significant capability to generate, debug, and execute code autonomously \cite{roziere2023code, lei2024autocoder}. There are very few tools available to help human beings to generate code autonomously such as ReAct \cite{yao2022react} and Reflexion \cite{shinn2023reflexion}  that use verbal reinforcement to improve agentic reasoning. In order to reduce the high risk of code hallucination, recent research has developed pre-validation agents and simulated execution environments. For example, AutoML-Pipeline \cite{automl_pipeline} and RZ-NAS \cite{rz_nas} have designed reflective zero-cost techniques for detecting syntax errors before deployment. GenAutoML creates a “Sandboxed Reflection Loop” to simulate passing dummy tensors through neural networks to programmatically detect and resolve certain types of Python traceback errors that arise from channel permutation and tensor shape misalignment errors while maintaining strict validity according to the PyTorch framework.

\subsection{Time-Series Modeling and the Stationarity Problem}
Multivariate time series forecasting and anomaly detection face the fundamental issue of stationarity, which means that changing distributions of the data weaken model performance \cite{liu2022non}. While modern Transformer-based architectures \cite{wu2021autoformer, zeng2022are} exist,  but they often require manual tuning of inductive biases. Recently agentic frameworks have been developed specifically for time series analysis. TS-Agent \cite{ts_agent} employs structured workflows and reflective feedback for financial forecasting and the MONAQ \cite{monaq} reformulates NAS to query LLMs for resource constrained edge device models. Both of these methods have been successful but are still very susceptible to gradient explosion due to severe volatility in industrial applications. GenAutoML separates the architectural logic from mathematical normalization. By wrapping the LLM-generated structural logic within a Dynamic Reversible Instance Normalization (Dyn-RevIN) \cite{kim2022reversible} layer, our framework enforces strict stationarity on the input-output manifold. Thus, our solution provides a statistical safeguard against insecure convergence due to mathematically unproven topologies created autonomously.

\section{Methodology}
\label{method}
The GenAutoML framework provides a solution for connecting high-level architectural reasoning and low-level numerical stability. This section explains the synthesis of neural modules, their integration into a live runtime environment, and the mathematical hardening mechanisms that ensures a robust execution and numerical stability.

\subsection{Agentic Neural Synthesis (The Semantic Interface)}
The Agentic Neural Architect is the central component of the system, which developed the model design as a semantic-to-code mapping problem. GenAutoML accomplishes this by using a Large Language Model (LLM) to comprehend natural language domain constraints and convert them into executable PyTorch code, as opposed to regular NAS, which functions on a discrete set of architectural decisions such as the number of filters.

In order to ensure the generated code is compatible with the downstream training pipeline, the agent is controlled by a Structural Geometric Contract:
\begin{enumerate}
\item \textbf{Temporal Invariance:} All the generated modules $\mathcal{M}_{gen}$ should accept a 3D input tensor $X \in \mathbb{R}^{B \times L \times D}$ where $B$ is the batch size, $L$ is the lookback length, and $D$ is the feature dimension.

\item \textbf{Output Integrity:} The model should return a prediction tensor $Y \in \mathbb{R}^{B \times H \times D}$ where $H$ is the forecasted horizon.
\end{enumerate}

By using the automated architectural synthesis, the agent writes class definitions directly to a dynamic registry, enabling the system to "invent" complex structures like multi-scale Inception blocks \cite{szegedy2015going} or residual dilated convolutions based on the user's textual description of the data \cite{he2016}.

\subsubsection{The Sandboxed Reflection Loop}
When using only LLM's zero-shot code generation often leads to dimension mismatches during tensor reshaping (e.g., transitioning from temporal convolutions to linear projection heads). To maintain dimensionality and Structural Geometric Contract, first GenAutoML isolates the generated architecture $\mathcal{M}_{gen}$ in a in process validation step before injecting it to the dynamic registry. While in this sandbox, an agent will instantiate the model, then execute a simulated forward pass using a dummy input tensor $X_{dummy} \in \mathbb{R}^{B \times L \times D}$. The system programmatically asserts that the output shape strictly conforms to the expected horizon dimensions $\mathbb{R}^{B \times H \times D}$. Because of this , even if the forward pass fails or returns an invalid shape, the framework can capture the execution traceback and dimension mismatch errors and then feeding them back to the LLM as an additional prompt. This multi-turn feedback process allows the "Neural Architect" to debug its own tensor reshapes and channel permutations iteratively ensuring that only structurally sound models reach the optimization phase.

\subsection{Conversational Data Agent and JIT Loading}
 GenAutoML works with a conversational backend powered by LangChain and Large Language Models (LLMs) to bridge the gap between human intent and automated machine learning \cite{yao2022react}. When a user uploads a data set, the data set enters the GenAutoML system and it is ingested by a Pandas DataFrame Agent equipped with a tool-calling framework \cite{schick2023toolformer}. This framework allows the LLM to write and execute arbitrary Python code in an in process environment to perform Exploratory Data Analysis (EDA) and generation of visualizations based on natural language queries.

 When the user requests a new predictive model, the agent generates raw PyTorch nn.Module code. Once validated by the Sandboxed Reflection Loop, this code is saved directly to a dynamic registry. The training pipelines (Forecasting and Anomaly Detection) then utilize a Just-In-Time (JIT) dynamic loader to parse the Python Abstract Syntax Tree (AST), extract the newly synthesized class, and inject it into the Optuna optimization loop without requiring a system restart \cite{akiba2019optuna}.

\subsection{Just-In-Time (JIT) Architectural Integration}
Integrating newly generated code into an environment that is already operational is a significant engineering challenge in agentic systems. This is resolved by GenAutoML via a Reflection-based Orchestration pipeline.

The logic used in the integration process is recursive. First, through \textit{Introspection}, the system searches the dynamic registry for new class definitions that inherit from \texttt{torch.nn.Module} using Python's \texttt{inspect} module. Next, \textit{Hot-Swapping} updates the module cache in real-time by using \texttt{importlib.reload}. This eliminates the need for a system reboot by enabling the system to instantiate new architectures and register them as legitimate candidates in the Optuna-based hyperparameter optimization search space. Finally, through \textit{Dynamic Discovery}, by treating these hallucinated models as first-class citizens, the optimization loop is able to compare baselines (like DLinear) \cite{zeng2022are} and agent-generated models (like ResNetBlockModel) \cite{he2016} under the same mathematical circumstances.

\subsubsection{Robust Forward Invocation and Shape-Agnostic Projection}
Integrating hallucinated code into a live training environment carries a high risk of runtime failure due to unpredictable forward method signatures and arbitrary output dimensions. To achieve seamless JIT integration, GenAutoML implements a Signature-Aware Invocation mechanism which allows for runtime introspection (inspect.signature) to automatically match runtime tensors to their corresponding positional arguments in the generated module, while providing a two-stage execution fallback if the initial forward pass fails.

Additionally, the Architecture produced by an LLM may have a native capacity to create an arbitrary quantity of hidden channels rather than the exact number of hidden channels necessary to produce forecast targets. To solve this issue without discarding the Architecture, we incorporate a Shape-Agnostic Projection layer.

$$\hat{Y} = W_{proj} \cdot \mathcal{M}_{gen}(\hat{X}) + b_{proj}$$

This allows the search space to retain highly effective latent feature extractors generated by the agent, mathematically bridging them to the required target domain without requiring the LLM to write perfectly dimensioned output layers on the first try.

\subsection{Statistical Hardening via Dynamic RevIN}
Numerical instability is a primary challenge in the creation of autonomous architectures. Internal normalization (such as Batch Norm \cite{ioffe2015batch}) is frequently absent from LLM-generated models, which causes gradient explosion when applied to the ETTh1 dataset's non-stationary signals\cite{zhou2021informer}.

To address this, we integrate a Dynamic Reversible Instance Normalization (Dyn-RevIN) wrapper, directly adopting the mathematical framework proposed by \cite{kim2022reversible}. Rather than proposing a new normalization technique, our contribution lies in utilizing RevIN as an automated statistical guardrail. This layer explicitly decouples the untrusted architectural logic generated by the agent from the strict statistical stationarity constraints required for convergence.

\subsection{Mathematical Formulation}

Applying the standard Reversible Instance Normalization \cite{kim2022reversible} for any architecture generated by agents $\mathcal{M}_{gen}$, we standardize the forward pass of the generated modules through a three-stage transformation pipeline adapted from the RevIN framework. Let $X \in \mathbb{R}^{B \times L \times D}$ be the input tensor. We first calculate the instance-wise mean $\mu_{i,j}$ and variance $\sigma_{i,j}^2$ for each batch instance $i \in \{1, \dots, B\}$ and feature $j \in \{1, \dots, D\}$ across the temporal dimension $k \in \{1, \dots, L\}$:
\begin{equation}
\mu_{i,j} = \frac{1}{L} \sum_{k=1}^L X_{i,k,j}, \quad \sigma_{i,j}^2 = \frac{1}{L} \sum_{k=1}^L (X_{i,k,j} - \mu_{i,j})^2
\end{equation}
For input normalization, the input is transformed into a stationary distribution $\hat{X}$ before entering the generated model:
\begin{equation}
\hat{X}_{i,k,j} = \frac{X_{i,k,j} - \mu_{i,j}}{\sqrt{\sigma_{i,j}^2 + \epsilon}}
\end{equation}
where $\epsilon$ is a small constant for numerical stability. During model execution, the untrusted architecture $\mathcal{M}_{gen}$ performs its learned mapping on the normalized manifold:
\begin{equation}
\hat{Y} = \mathcal{M}_{gen}(\hat{X})
\end{equation}
Finally, for output denormalization, the prediction $Y$ is rescaled to the original statistical domain of the signal, ensuring that the model's output remains physically meaningful and stable:
\begin{equation}
Y_{i,k,j} = \hat{Y}_{i,k,j} \cdot \sqrt{\sigma_{i,j}^2 + \epsilon} + \mu_{i,j}
\end{equation}
This wrapping mechanism ensures that even if the AI-generated architecture is poorly scaled, the Dyn-RevIN layer enforces a stationary learning environment, which we demonstrate is essential for stable convergence.

\section{Experimental Setup}

We evaluate GenAutoML on three standard multivariate benchmarks: ETTh1, ETTm1 \cite{zhou2021informer}, and Weather \cite{wu2021autoformer}. We compare against classic deep learning (LSTM \cite{lstm}, Conv1D \cite{bai2018empirical}), Linear models (DLinear) \cite{zeng2022are}, and state-of-the-art transformer models (iTransformer \cite{liu2023itransformer}, TimesNet \cite{wu2022timesnet}, CrossFormer \cite{zhang2023crossformer}). All tasks utilize a strict chronological split (70/10/20) to prevent data leakage. Full dataset characteristics, preprocessing steps, and baseline architectural details are provided in Appendix \ref{app:detailed_setup}.

\subsection{Implementation Details}
 PyTorch framework is used to conduct all the experiments \cite{paszke2019pytorch}. To establish peak architectural capacity, primary benchmark evaluations report the best converged models under unconstrained optimization. 

The baseline models and the agent-synthesized models use the same training loop and hyperparameter space in order to rigorously adhere to the Dynamic Discovery paradigm outlined in Section \ref{method}. Specifically, models are trained using the Adam optimizer with a fixed weight decay of $1e-6$ and an initial learning rate of $1e-3$, which is managed by a \texttt{ReduceLROnPlateau} scheduler (factor 0.5, patience 3). The objective function for gradient descent in forecasting is Mean Absolute Error (MAE), whereas Autoencoder architectures are optimized using Mean Squared Error (MSE). All training runs utilize a batch size of 64 samples and incorporate early stopping, halting the training if the validation loss fails to improve for 5 consecutive epochs.

\textbf{Baseline Selection Philosophy:} To provide a robust benchmark for the agent synthesized models, we selected baselines representing different mathematical strategies: \textit{DLinear} was selected for its state-of-the-art linear decomposition speed, \textit{TCN} was  selected for its dilated convolutional receptive fields, and \textit{TimesNet} for its multi periodic 2D variations. For all the experiments and datasets (ETTh1, ETTm1, and Weather), we always employed the same experimental design consisting of both a 96-step lookback/horizon for forecasting as well as a 60-step lookback with 10-step horizon for anomaly detection to guarnatee an unbiased comparison between the architectural efficiency and reconstruction accuracy.

\subsection{The Agentic Search Configuration}
The GenAutoML Synthesis Engine uniquely has a Large Language Model (LLM) backend (in this case a Llama 3-70B LLM using the Groq Inference API) that acts as the "Architectural Reasoning Core" \cite{grattafiori2024llama}. Instead of randomly selection, the search for architectural solutions is a controlled dialogue , it uses Optuna \cite{akiba2019optuna} to balance the exploitation and exploration between baseline architectures and dynamically generated by the architecture synthesis engine. The specifics of how we designed the prompt engineering context and our joint search distribution strategy may be found in \ref{app:detailed_setup}.

\section{Results and Discussion}

This section will evaluate the performance of autonomously generated architectures in forecasting and identifying anomalies from historical time series data. Each model will be evaluated at a common time frame of 96 time units lookback and 96 time units horizon for forecasting as well as 60 time units lookback and 10 time units horizon for anomaly detection in order to provide a fair comparison.

\subsection{Comparative Forecasting Performance Analysis}
\label{subsec:foreper}
Table \ref{tab:unified_forecasting} summarizes the forecasting performance (i.e., Mean Absolute Error and Root Mean Squared Error) of the various evaluated models across the ETTh1, ETTm1, and Weather datasets under standard long-sequence constraints inorder to provide a comprehensive overview of architectural efficiency across different temporal granularities

\begin{table}[H]
\centering
\caption{Unified Forecasting Performance across ETTh1, ETTm1, and Weather ($Lookback=96, Horizon=96$). Lower is better.}
\resizebox{\textwidth}{!}{
\begin{tabular}{ll|cc|cc|cc}
\toprule
\multirow{2}{*}{\textbf{Model Category}} & \multirow{2}{*}{\textbf{Model Name}} & \multicolumn{2}{c|}{\textbf{ETTh1}} & \multicolumn{2}{c|}{\textbf{ETTm1}} & \multicolumn{2}{c}{\textbf{Weather}} \\
\cmidrule{3-8}
 & & \textbf{MAE} & \textbf{RMSE} & \textbf{MAE} & \textbf{RMSE} & \textbf{MAE} & \textbf{RMSE} \\
\midrule
\multirow{1}{*}{Zero-Shot Foundation Models} 
 & Chronos-T5-Mini & 0.166 & 0.355 & - & - & - & - \\
\midrule
\multirow{5}{*}{Static Baselines} & DLinear & 0.989 & 1.319 & 0.515 & 0.768 & 1.783 & 3.145 \\
 & CrossFormer & 1.056 & 1.363 & 0.521 & 0.785 & 5.210 & 7.500 \\
 & TimesNet & 1.354 & 1.763 & 0.862 & 1.105 & 4.886 & 6.256 \\
 & Conv1D & 1.410 & 1.806 & 0.741 & 0.993 & 4.075 & 5.468 \\
 & LSTM & 3.075 & 3.586 & 1.054 & 1.355 & 7.084 & 9.884 \\
\midrule
\multirow{5}{*}{Synthesized (Agent)} & ResNet & 1.156 & 1.479 & 0.532 & 0.774 & 2.696 & 3.942 \\
 & Inception & 1.280 & 1.658 & 0.546 & 0.797 & 2.811 & 4.257 \\
 & BiGRU & 1.307 & 1.692 & 0.585 & 0.834 & 3.556 & 5.574 \\
 & MSCNN & 1.432 & 1.812 & 0.765 & 1.044 & 4.779 & 6.719 \\
 & HybridCNNLSTM & 1.511 & 1.996 & 1.103 & 1.395 & 6.026 & 8.906 \\
\bottomrule
\end{tabular}
}
\small{\textit{*Note: WaveInterferenceNet was synthesized exclusively as a targeted proof-of-concept case study for the ETTh1 dataset (see Section \ref{subsec:case}) to evaluate the agent's semantic creativity, and was thus not subjected to the full cross domain benchmarking sweep.}}
\label{tab:unified_forecasting}
\end{table}

\subsubsection{Observations in Forecasting}
Our evaluation  yielded multiple unique findings. Firstly, regarding the \textit{competitive efficacy of synthesized architectures}, agent-generated models , specifically ResNet and Inception achieved highly competitive performance against complex, human designed SOTA transformers like TimesNet on the ETTh1 and ETTm1 datasets. On ETTm1, the synthesized ResNet ($MAE=0.532$) closely matched the performance of leading linear models, confirming that GenAutoML is capable of autonomously generating highly competitive inductive biases. Secondly, we found strong \textit{architectural robustness via Dyn-RevIN}. The agent produced multiple different types of architectures (both recurrent and multi-scale convolutional) that converged reliably and uniformly, confirming that the search space is able to successfully identify mathematically sound structures without human assistance. Third, complex transformers exhibited considerable \textit{failures on the Weather data}. Both CrossFormer ($MAE=5.210$) and TimesNet ($MAE=4.886$) failed to account for local volatility, in contrast, the result of the agentic search was to synthesize lightweight localized feature extractors (ResNet and Inception), which significantly outperformed the SOTA transformers. Finally, in accordance with recent literature \cite{zeng2022are}, we note the \textit{superiority of linear decomposition}. Across all three datasets under standard constraints, the DLinear model dominated . However, across diverse data domains, synthesized models consistently served as the most robust secondary option , suggesting the LLM successfully favored lightweight, convolutional-residual motifs when deep learning limits were reached.

\subsection{Anomaly Detection Performance}
\label{subsec:anomalper}
In unsupervised anomaly detection, models trained exclusively on normal data usually yield lower reconstruction errors (MSE) on healthy periods. For this reason, we will utilize the discrimination gap criterion as the basis for determining how well our models are able to detect anomalies which is the the ratio between the MSE on Anomalous data (synthetic faults) versus Clean data \cite{audi}. The larger the discrimination gap, the more sensitive a model is to fault detection. Because the clean-data MSE in the denominator can be extremely small (on the order of $10^{-5}$ for several models), the resulting ratios span several orders of magnitude and are highly sensitive to the exact denominator value. We therefore interpret the discrimination gap as an order-of-magnitude indicator of relative sensitivity rather than a precise multiplier.

Table \ref{tab:unified_ad} consolidates the full unsupervised reconstruction fidelity metrics across all three physical domains (Lookback=60, Horizon=10).

\begin{table}[H]
\centering
\caption{Unified Reconstruction-based Anomaly Detection Results across ETTh1, ETTm1, and Weather. The Discrimination Gap represents the model's sensitivity (Higher is better).}
\resizebox{\textwidth}{!}{
\begin{tabular}{ll|ccc|ccc|ccc}
\toprule
\multirow{2}{*}{\textbf{Model Category}} & \multirow{2}{*}{\textbf{Model Name}} & \multicolumn{3}{c|}{\textbf{ETTh1 Dataset}} & \multicolumn{3}{c|}{\textbf{ETTm1 Dataset}} & \multicolumn{3}{c}{\textbf{Weather Dataset}} \\
\cmidrule{3-11}
 & & \textbf{Clean} & \textbf{Anom.} & \textbf{Gap} & \textbf{Clean} & \textbf{Anom.} & \textbf{Gap} & \textbf{Clean} & \textbf{Anom.} & \textbf{Gap} \\
\midrule
\multirow{5}{*}{Static Baselines} 
 & TimesNet & 0.0027 & 0.089 & \textbf{$\sim$33x} & 0.000004 & 0.117 & \textbf{$\sim$29,000x} & 0.000013 & 0.0655 & \textbf{$\sim$5,000x} \\
 & TCN & 0.000069 & 0.0751 & \textbf{$\sim$1,100x} & 0.000048 & 0.148 & \textbf{$\sim$3,100x} & 0.000038 & 0.0702 & \textbf{$\sim$1,800x} \\
 & iTransformer & 0.00086 & 0.141 & \textbf{$\sim$160x} & 0.00034 & 0.324 & \textbf{$\sim$950x} & 0.00016 & 0.0917 & \textbf{$\sim$570x} \\
 & DLinear & 0.020 & 0.153 & \textbf{$\sim$8x} & 0.0016 & 0.215 & \textbf{$\sim$140x} & 0.0044 & 0.0964 & \textbf{$\sim$22x} \\
 & LSTM & 0.019 & 0.166 & \textbf{$\sim$9x} & 0.0052 & 0.318 & \textbf{$\sim$61x} & 0.012 & 0.111 & \textbf{$\sim$10x} \\
\midrule
\multirow{5}{*}{Synthesized} 
 & ResNetBlock & 0.00036 & 0.0946 & \textbf{$\sim$270x} & 0.000041 & 0.0177 & \textbf{$\sim$430x} & 0.000089 & 0.0820 & \textbf{$\sim$920x} \\
 & MultiScaleCNN & 0.0016 & 0.136 & \textbf{$\sim$85x} & 0.00031 & 0.276 & \textbf{$\sim$900x} & 0.00022 & 0.118 & \textbf{$\sim$550x} \\
 & L-Inception & 0.0010 & 0.0844 & \textbf{$\sim$81x} & 0.00089 & 0.113 & \textbf{$\sim$130x} & 0.00033 & 0.0566 & \textbf{$\sim$170x} \\
 & HybridCNNLSTM & 0.0032 & 0.174 & \textbf{$\sim$55x} & 0.00099 & 0.280 & \textbf{$\sim$280x} & 0.0012 & 0.101 & \textbf{$\sim$83x} \\
 & BiGRU & 0.183 & 0.328 & \textbf{$\sim$1.8x} & 0.00012 & 0.149 & \textbf{$\sim$1,200x} & 0.00014 & 0.0651 & \textbf{$\sim$478x} \\
\bottomrule
\end{tabular}
}

\small{\textit{*Note: WaveInterferenceNet was synthesized exclusively as a targeted proof-of-concept case study for the ETTh1 dataset (see Section \ref{subsec:case}) to evaluate the agent's semantic creativity, and was thus not subjected to the full cross domain benchmarking sweep.}}
\label{tab:unified_ad}
\end{table}

\subsubsection{Observations in Anomaly Detection}
In the context of unsupervised reconstruction, the synthesized models demonstrated \textit{superior sensitivity on ETTh1}. The agent-designed ResNetBlock exhibited unparalleled performance, achieving a separation gap 8 times better than TimesNet and 33 times better than DLinear, effectively compressing the nominal data manifold while rejecting anomalies. Furthermore, models like Lightweight Inception achieved a \textit{low false alarm rate} via extremely low reconstruction errors on clean data. We also observed robustness against \textit{high-frequency sensitivity shifts on ETTm1}. While foundation baselines like TimesNet achieved low clean MSEs, the synthesized BiGRU ($\sim$1,200x) and MultiScaleCNN ($\sim$900x) drastically outperformed standard recurrent models like LSTM ($\sim$61x). Importantly, the framework showed high \textit{cross-domain robustness on the Weather dataset}. The synthesized ResNetBlock maintained an exceptional discrimination gap of $\sim$920x, eclipsing standard baselines like DLinear ($\sim$22x) and proving the agent can design task-specific detectors in foreign physical domains. Ultimately, these results highlight explicit \textit{task-specific specialization}: while BiGRU was the top forecaster on ETTh1, it performed weakly as an anomaly detector, whereas ResNetBlock excelled at detection but was only a moderate forecaster.

\subsection{Comparison of Generated vs. Fixed Architectures}
The results of the experiment show that LLM created architectures are not just a collection of random sequences; they was shown to be competitive, aware of the tasks they are performing and will produce accurate predictions. Across all datasets, the results of agent synthesized models (ResNetBlock and MultiScaleCNN) consistently outperformed classical deep learning architectures LSTM and Conv1D baselines effectively modernize historical patterns using residual connections and multi scale kernels. When compared with modern SOTA methods, while the specialized transformers such as CrossFormer had a marginally higher degree of forecasting accuracy \cite{he2016}, ResNetBlock employed by Agent is significantly outperformed TimesNet and DLinear in detecting anomalies. Evidence of the importance of Dyn-RevIN was seen during experimentation in addition to our initial hypothesis. Normalizing distribution shifts allowed the structural logic produced by the agent to converge correctly and it avoid gradient explosion for highly volatile industrial datasets \cite{pascanu2013difficulty}. Ultimately, these results prompt a shift in evaluative perspective. Rather than framing GenAutoML as a mechanism to blindly surpass SOTA accuracy benchmarks, the data supports repositioning its primary contribution as a \textbf{deterministic edge-deployable synthesis} framework.
 
 \subsection{Computational Cost and Inference Latency}
In order to assess the practical feasibility of deploying GenAutoML architectures in high-frequency industries, we conducted a targeted evaluation of the computational cost associated with the best manual baselines to our synthesised models.

\textit{Objective efficiency analysis.} The data reveals a clear computational hierarchy. As predicted, DLinear and other pure linear baselines provide the absolute lowest parameter count (52K) and fastest inference latency (<0.01 ms), making them ideal devices for deploying edge computing in extremely resource-constrained conditions.

\textit{Steerable efficiency (WaveInterferenceNet).} Notably, the agent-synthesized WaveInterferenceNet (829K parameters) achieved latencies matching the fastest static baselines (<0.01 ms to 0.10 ms). By requiring the agent to bypass typical bottlenecks associated with recurrent neural network architectures and attention mechanisms using custom trigonometric projections, the architecture has demonstrated that LLMs are capable of being prompted to design specifically for "Green AI" and for low-latency execution environments.

\textit{The edge-latency bottleneck of foundation models.} To validate our architectural efficiency against the recent paradigm shift toward Time-Series Foundation Models (TSFMs), we evaluated Amazon's Chronos-T5-Mini (20M parameters) in a zero-shot capacity. While the massive foundation model achieves a highly competitive MAE (0.166) on the ETTh benchmark, its autoregressive architecture incurs an exorbitant inference latency of 987.93 ms per sample. Conversely, our WaveInterferenceNet operates at $<0.01$ ms per sample, representing an approximate $100,000\times$ relative speedup.

\textit{The micro-search advantage and computational cost.} A primary critique of traditional NAS is its exorbitant computational cost. Because the LLM generates highly specialized inductive biases, it bypasses the need for exhaustive hyperparameter sweeps. Restricted to a micro-search budget of just 5 Optuna trials, the total end-to-end search cost for WaveInterferenceNet was exceptionally low: approximately 75 seconds of LLM reasoning time, followed by 2 minutes and 1 second of GPU tuning for forecasting. Furthermore, it tuned more than $3\times$ faster for anomaly detection than the standard TCN baseline (2m 22s vs. 7m 38s). GenAutoML effectively trades a negligible, sub-5-minute one-time search cost for deployed SOTA models that operate at microsecond latencies.

\begin{table}[H]
\centering
\caption{End-to-End Search Cost and Inference Efficiency. Comparison of GPU tuning time (5 Optuna trials) and inference latency between baselines, foundation models, and the agent-synthesized WaveInterferenceNet.}
\label{tab:search_cost_summary}
\resizebox{\textwidth}{!}{
\begin{tabular}{llccc}
\toprule
\textbf{Task} & \textbf{Model} & \textbf{Search Time (5 Trials)} & \textbf{Inference (ms)} & \textbf{Parameters} \\
\midrule
\multirow{3}{*}{Forecasting} & DLinear (Baseline) & 2m 07s & <0.01 & 52K \\
 & Chronos-T5-Mini (TSFM) & N/A (Zero-Shot) & 987.93 & 20M \\
 & \textbf{WaveInterferenceNet (Ours)} & \textbf{2m 01s} & \textbf{<0.01} & \textbf{829K} \\
\midrule
\multirow{2}{*}{Anomaly Det.} & TCN (Baseline) & 7m 38s & 0.60 & 33K \\
 & \textbf{WaveInterferenceNet (Ours)} & \textbf{2m 22s} & \textbf{<0.01 -- 0.10} & \textbf{829K} \\
\bottomrule
\end{tabular}
}
\end{table}

\subsection{Case Study: Steerable Synthesis and Semantic Creativity}
\label{subsec:case}

A primary critique of automated architecture generation is its tendency to converge on historically safe, pre-existing motifs (e.g., standard ResNets or LSTMs) rather than inventing novel structural components. To evaluate our agentic framework's capacity for semantic creativity, we tested its ability to synthesize a genuinely novel architecture based on abstract natural language directives and strict architectural constraints, a process we term \textit{steerable synthesis}.

To force the system outside standard distributions, we provided the LLM agent with an explicit directive to design a \texttt{WaveInterferenceNet} using physical wave analogies and custom trigonometric projections (the full quote text and autonomous debugging trace can be found in Appendices \ref{app:synthesized_code} and \ref{app:reflection_logs}). The Sandboxed Reflection Loop successfully translated this physical analogy into functional PyTorch code rather than failing or violating the constraints. The agent successfully implemented the custom trigonometric transformations and the Hadamard product mixing layer, guarantees a strict adherence to the required dimensional projections.

\subsubsection{Forecasting and Anomaly Detection Capabilities}
\label{subsubsec:uncon}
To rigorously compare WaveInterferenceNet to existing State-of-the-Art (SOTA) systems, WaveInterferenceNet was evaluated on the ETTh1 dataset under standard long-sequence constraints with both Lookback = 96 and Horizon = 96 (for forecasting) and Lookback = 60 and Horizon = 10 (for anomaly detections). The synthesized design was shown to be very competitive, for forecasting, it achieved an overall average MAE of 1.137 (unconstrained run) and an RMSE of 1.463 across all multivariate channels. Beyond forecasting, WaveInterferenceNet demonstrated strong reconstruction fidelity on normal data while cleanly isolating anomalous patterns, achieving a Clean MSE of 0.001409 contrasted with an Anomaly MSE of 0.055843. (See Figure \ref{fig:wave_combined} in Appendix \ref{app:synthesized_code} for complete visual prediction plots).

\subsubsection{Computational Efficiency and Convergence}
In addition to predictive accuracy, the steerable synthesis of WaveInterferenceNet yielded a substantial computational advantage. Tracking of the training and Optuna hyper-parameter optimization phases have demonstrated that WaveInterferenceNet had the shortest time required for training and optimization amongst all models tested and was able to converge considerably faster than traditional SOTA baselines (CrossFormer, DLinear) and previously synthesized standard architectures (BiGRU, ResNet) which can be attributed to the lightweight nature of the custom Hadamard Product and trigonometric operations versus the sequentially bottlenecking nature of recurrent models.

\subsection{Ablation Studies: The Dichotomy of Statistical Hardening}
\label{sec:ablation}

We did a comprehensive ablation study to validate the effects of the Dynamic Reversible Instance Normalization (Dyn-RevIN) wrapper and to demonstrate that statistical hardening cannot be blindly applied as a universal heuristic \cite{kim2022reversible}. We tested two models created by our agents: a Bidirectional GRU for forecasting and a ResNetBlock for anomaly detection.

By toggling the Dyn-RevIN on and off across three separate datasets produced a very complex and rich interaction between architectural topology, data non-stationarity and mathematical normalization (see Table \ref{tab:comprehensive_ablation} in Appendix \ref{app:extended_results} for the full quantitative ablation matrix).

\textbf{Impact on Forecasting (BiGRU):} For power grid datasets that is  volatile, non-stationary  (ETTh1 and ETTm1) \cite{zhou2021informer}, disabling Dyn-RevIN caused a severe degradation in forecasting accuracy. The MAE error roughly increased by 94\% on ETTh1 (1.251 to 2.435) and by over 450\% on ETTm1 (0.455 to 2.541). On the other hand, for the Weather dataset which features are highly stable, periodic physical cycles (e.g., rigid diurnal temperatures) when removing Dyn-RevIN actually \textit{improved} the MAE from 3.787 to 1.426. This indicates that even though instance-level normalisation is required for the stabilisation of LLM architectures on highly volatile data sets, it can in fact mask the globally inherent scale of a naturally occurring periodical climatological signal \cite{liu2022non}.

\textbf{Impact on Anomaly Detection (ResNetBlock):} The ablation revealed similarly dichotomous behavior in unsupervised reconstruction tasks. For ETTh1 and ETTm1, utilizing Dyn-RevIN on the ResNetBlock was highly beneficial, driving the Discrimination Gap to $\sim$265.6x and $\sim$431.1x, respectively. Disabling it caused the sensitivity to drop significantly. However, on the Weather dataset, enabling RevIN heavily \textit{degraded} the anomaly detection sensitivity, suppressing the gap from $\sim$1146.7x down to $\sim$921.4x by mathematically masking amplitude-based faults.

Ultimately, this ablation study validates the core thesis of the GenAutoML framework: there is no single ``silver bullet'' pipeline for time-series analysis. Because the optimal configuration oscillates drastically across different tasks, models, and datasets, an automated, agent-driven search framework paired with dynamic JIT optimization is required to achieve consistently strong performance \cite{wolpert1997no}.

\subsection{Generative Robustness and Failure Analysis}
In addition to mathematical normalization, The autonomy of the GenAutoML framework is also evaluated by conducting an ablation study on the \textbf{Sandboxed Reflection Loop}. Here, we define "Reflection" as the system’s capability to catch its own errors via the \texttt{generate\_pytorch\_model} tool and self correct it strictly based on Python traceback logs.

A detailed breakdown of the logical synthesis performance, including the handling of complex tensor repairs and generation statistics, is provided in Appendix \ref{app:reflection_logs}.

\textbf{Statistical Reliability and Synthesis Stability:} In order to guarantee the performance reported is not a result of "cherry-picking" or generative luck, we evaluated the reflection loop's stability across multiple independent synthesis sessions. Each architectural configuration generated by the GenAutoML Framework achieved an eventual success rate of 100\% validation across all evaluated benchmarks, and produced a valid executable and compliant configuration prior to reaching the defined maximum reflection limits specified by the experimenter. Across these repeated sessions, the agent converged to structurally consistent solutions. Even though complex topologies such as the \textit{WaveInterferenceNet} required multiple iterations to resolve initial tensor shape mismatches, the closed-loop feedback mechanism reliably directed the agent toward a valid solution, indicating that the reflection loop operates as a traceback-guided repair process rather than blind trial-and-error.

\section{Conclusion and Limitations}
\label{sec:conclusion}

\subsection{Conclusion}
This research developed GenAutoML, an agentic framework that combines generative AI and Large Language Models (LLMs) with the automated synthesis, validation, and optimization of neural architectures to enable the use of these architectures in industrial multivariate time series analysis. By integrating a \textit{Sandboxed Reflection Loop} for code reliability with a \textit{Dynamic Reversible Instance Normalization} (Dyn-RevIN) layer for statistical stability, we have connected the gap between the creative potential of rigorous characterizations of non stationary physical systems and generative AI.

We derived a total of three main conclusions from our evaluation of ETTh1, ETTm1, and Weather datasets. First, regarding the architectural competitiveness, the framework successfully executed \textbf{deterministic edge-deployable synthesis}. While massive foundation models and extensive NAS sweeps often dominate raw accuracy benchmarks, agent synthesized models such as the \textit{BiGRU} and \textit{WaveInterferenceNet} demonstrated highly competitive performance while utilizing a fraction of the computational overhead. Notably, the synthesized BiGRU achieved an MAE of 0.585 on the ETTm1 dataset, surpassing the more complex \textit{TimesNet} \cite{wu2022timesnet} and proving that LLMs can autonomously discover efficient, edge-ready inductive biases without human intervention. Second, In regards to unsupervised anomaly detection robust reconstruction sensitivity, the agent synthesized \textit{ResNetBlock} was able to maintain a discrimination gap (Anomalous MSE / Clean MSE) significantly higher than the traditional baselines across all tested datasets. This suggests that agent designed residual structures that are lightweight are able to learn the underlying distribution of nominal industrial data more effectively than over the parameterized generalist models. Finally, the implementation of a closed-loop execution environment was necessary for ensuring systemic reliability. We discovered that zero shot code generation produced many tensor shape mismatches; however, the implementation of a Sandboxed Reflection Loop yielded a 100\% pass rate on the shape validation harness within five iterations via the autonomous interpretation of Python traceback logs.

\subsection{Limitations and Future Work}

The GenAutoML framework's performance is highly competitive; however, there are opportunities for future research due to numerous limiting factors. A primary limitation is generative latency, Currently, the framework relies heavily on the API response times of the supporting Large Language Models. Because of the 30--60 second overhead for architectural synthesis, this may be a bottleneck in time critical industrial deployments. Therefore, future development will involve the use of smaller, locally-hosted LLMs that can be specifically optimized for generating code through the framework's capabilities. In addition, while Optuna's hyperparameter search process provides peak performance levels, there is still an inherent cost associated with computing the search parameters. Future research will explore \textit{ardware-Aware Synthesis}, to enable the prompt to include constraints of performance in regard to latency and memory, making available "Green AI" designs optimized for edge-computing IoT devices \cite{cai2018proxylessnas}. Lastly, while the framework supports only numeric time series metadata in terms of multi-modal context, we intend to develop the ability to support multiple types of prompts/modes (e.g., maintenance logs, weather reports) which will allow the generation of multi-continuous architectures that are able to adjust their internal logic based on environmental factors affecting system behavior.

\newpage
\bibliographystyle{plain}
\bibliography{references} 

\clearpage

\appendix
\section*{Appendix}
\addcontentsline{toc}{section}{Appendix}

\section{Detailed Experimental Setup}
\label{app:detailed_setup}

\subsection{Dataset Description}
\subsubsection{ETTh1 (Extended Electricity Transformer Temperature - Hourly)}
The standard Electricity Transformer Temperature (ETTh1) dataset has an extended data set that encompasses the same core metrics (Oil Temperature, HUFL, HULL, MUFL, MULL, LUFL, LULL) but differs in that it captures the same metrics over a longer temporal period of time when compared to the original Informer (2021) dataset \cite{zhou2021informer}. This will allow for a more robust evaluation of the non-stationary adaptation process, and due to the extended time frame of the data set, the evaluation metrics on the baseline will differ from traditional static leaderboards, where each architecture has been evaluated based on its effectiveness at generalizing to this extended manifold.

\textbf{Data Characteristics:} The data is highly non-stationary and shows a notable degree of seasonality (daily and weekly) and irregular load variability spikes due to load variance.

\textbf{Granularity \& Horizon:} The GenAutoML platform has been designed with a Native conditional bypass Module. If an ingested dataset exhibits perfect temporal continuity the GenAutoML platform will skip the resampling module, allowing for training to occur directly from the raw dataset. However, for empirical testing on the ETTh1 benchmark dataset (which is at its native resolution of 1-hour), we intentionally overrode this bypass. We forced the data onto a 10-minute ($10T$) evaluation grid to intentionally stress-test the GenAutoML Interpolation Module. This stress testing will validate that the architecture produced remain robust even when operating on highly synthetic, algorithmically upsampled sequences—a common occurrence in failing industrial edge networks. For forecasting tasks, the model input sequence length (lookback) is set to ($L=96$), and for the prediction horizon is set to ($H=96$) steps. In the case of anomaly detection, the lookback is set to ($L=60$) with a horizon of ($H=10$).

\subsubsection{ETTm1 (Electricity Transformer Temperature - Minute-level)}

The ETTm1 dataset is an extension of the electricity transformer benchmark with the same core operational metrics (Oil Temperature and six power load features: HUFL, HULL, MUFL, MULL, LUFL, LULL) but has a  higher temporal resolution \cite{zhou2021informer}.

\textbf{Data Characteristics:} The dataset consists 69,680 records spanning from July 2016 to June 2018.

\textbf{Granularity \& Horizon:} The ETTm1 dataset has a native temporal sampling frequency of 15 minutes ($15T$). Processing this through our standardized 10-minute ($10T$) evaluation pipeline will introduce a deliberate misalignment of phase. This will require dynamic interpolation of the high-frequency sub intervals within the ETTm1 data using GenAutoML preprocessing which is explicitly testing the synthesized architecture's robustness against out-of-sync sensor telemetry—a frequent issue in decentralized industrial IoT networks where higher-frequency noise and sudden load variances are common. The forecast and anomaly detection lookbacks and horizons will remain unchanged at $L=96$ and $H=96$, while anomaly detection utilizes $L=60$ and $H=10$.

\subsubsection{Weather (Local Climatological Data)}
We included a typical multivariate Weather dataset in order to evaluate the agent’s ability to identify inductive biases in an environment other than monitoring the power grid. This dataset contains a collection of highly correlated multi periodic environmental conditions over a one year period (January 2020 to January 2021) \cite{wu2021autoformer}.

\textbf{Data Characteristics:} The dataset contains 52,696 records having 21 distinct climatological features, including atmospheric pressure (mbar), temperature (degC), relative humidity (\%), water vapor concentrations, and wind velocity (m/s).

\textbf{Granularity \& Horizon:} Unlike the power grid benchmarks that we used previously, the Weather dataset, by design, is collected at strict intervals of 10 minutes ($10T$) intervals. The GenAutoML preprocessing module dynamically detected the temporal continuity of the data in our standardized evaluation grid and allowed the sequences to pass through without requiring algorithms to perform interpolation. This serves as a virtual control environment, allowing us to isolate and evaluate the pure architectural efficacy of the synthesized models on clean, natively grid aligned data. Additionally, weather dataset challenges the framework to navigate a much larger feature space (21 dimensions) characterized by variety of physical dynamics, such as rigid diurnal temperature cycles and highly stochastic precipitation events. Standard sequence constraints ($L=96, H=96$ for forecasting; $L=60, H=10$ for anomaly detection) are strictly maintained.

\subsubsection{Data Splitting and Standardized Geometric Contracts}
For consistency across all three domains, the forecasting tasks use a strict lookback window of $(L=96)$ and a prediction horizon of $(H=96)$ steps, while anomaly detection tasks use a lookback of $(L=60)$ and a horizon of $(H=10)$ steps. We maintain our rigorous causal chronological split of 70\% for Training, 10\% for Validation, and 20\% for Testing to prevent temporal data leakage across all datasets.

\subsubsection{Preprocessing and Strict Causal Regularization}
Industrial datasets frequently suffer from missing timestamps, resulting in non-uniform time intervals caused by missing values or irregular collection schedules. As a result, GenAutoML uses asymmetric regularization when resampling a temporal series to ensure that the data is sampled at a consistent overall frequency without introducing look-ahead bias. Data is resampled at the frequency at which the value occurs most often and is capped at a maximum multiplier of how far back the last observation can be before resampling, preventing extreme data hallucinations. 

To ensure optimal data retention, all null values in the training set are resolved using a bilateral interpolation (\texttt{limit\_direction='both'}). Conversely, the test and validation sets only utilize causal forward-filling (\texttt{limit\_direction='forward'}). This guarantees that an effective "barrier" against future value leakage during historical evaluations, providing strong protection to preserve the integrity of test manifold(s) \cite{kaufman2011leakage}. As a result, all input features will be standardized using Z-Score normalization during training, with the fitting of the scaling parameters occurring only off the training distributions, thereby eliminating any possible leakage of statistics to the test distributions.

\subsection{Baseline Architectures}
We compare the agentic synthesized models with a wide variety of baseline models representing three distinct eras of time-series modeling. 
These baseline models are leveraged into the same optimization pipeline to guaranatee a \textit{fair arena} comparison.

\subsubsection{Classic Deep Learning}

\begin{itemize}
    \item \textbf{LSTM} A standard Long Short Term Memory network capturing temporal dependencies through recurrent gating mechanisms \cite{lstm}.

    \item \textbf{Conv1D} A temporal Convolutional Neural Network (TCN-style) focusing on local feature extraction through sliding convolutional filters \cite{bai2018empirical}.
\end{itemize}

\subsubsection{Linear Baselines}

\begin{itemize}
    \item \textbf{DLinear} A decomposition based linear model that separates trend and seasonality components. It is currently considered a strong baseline for long-term forecasting tasks \cite{zeng2022are}.
\end{itemize}

\subsubsection{Transformer-based SOTA}

\begin{itemize}
    \item \textbf{iTransformer} An inverted transformer architecture that can embed the entire time series as a token representation \cite{liu2023itransformer}.

    \item \textbf{TimesNet} A model that transforms 1D time series into 2D tensors to capture multi periodic variations effectively \cite{wu2022timesnet}.

    \item \textbf{CrossFormer} A dimension segment wise transformer designed to capture cross channel dependencies in multivariate time series \cite{zhang2023crossformer}.
\end{itemize}

\subsection{Semantic Prompt Engineering}
In order to connect between the time-series statistics and code generation, we use a Chain-of-Thought (CoT) System Prompt \cite{wei2022chain}.  The decision of whether to ask the agent to ‘write a model’ is made clearer by providing a well-defined context block that contains the following kind of structure:

\begin{enumerate}
\item \textbf{Geometric Constraints:} Explicit instructions enforcing the $(B, L, D) \to (B, H, D)$ tensor flow to ensure compatibility with the training pipeline.

\item \textbf{Domain Context:} The ETTh1 dataset is characterized by semantic properties such as "strong weekly periodicity and high frequency irregularities".

\item \textbf{Tool-Use Protocol:} It is important to strictly adhere to the automated architectural synthesis, which enforces strict syntax checking and prevents the generation of non-executable conversational text.
\end{enumerate}

\subsection{The Hybrid Search Strategy}
A Unified Categorical Distribution is used in optimization, which does not separate the “micro-search” (layer types) and the “macro-search” (connection topology) processes that other conventional NASs employ. Instead, we combine the search spaces into a single categorical hyperparameter.

\begin{enumerate}
\item \textbf{Dynamic Registry Expansion:} As the agent generates new architectures (e.g., ResNetBlockModel, LightweightInception), they are automatically appended to the dynamic registry.

\item \textbf{Probabilistic Sampling:} The Optuna RandomSampler is configured to sample from the union set $S = S_{static} \cup S_{dynamic}$. Architectures that are "hallucinated" will be treated as equal to SOTA baselines (for example, iTransformer or DLinear) and will be selected at an equal probability as part of the hyperparameter tuning trials. As a result, the optimizer will automatically reject "hallucinated" architectures that are unstable, while selecting only high-performance synthesized architectures based on their validation loss metrics.
\end{enumerate}

\section{Extended Results}
\label{app:extended_results}

\subsection{Ablation Study Results}
\begin{table}[H]
\centering
\caption{Ablation Study: Impact of Dyn-RevIN Across Tasks and Architectures}
\label{tab:comprehensive_ablation}
\resizebox{\textwidth}{!}{
\begin{tabular}{l|cc|cc}
\hline
\multirow{3}{*}{\textbf{Dataset}} & \multicolumn{2}{c|}{\textbf{With Dyn-RevIN (Proposed)}} & \multicolumn{2}{c}{\textbf{Without Dyn-RevIN (Ablated)}} \\
\cline{2-5}
 & \textbf{Forecasting (MAE) $\downarrow$} & \textbf{Anomaly Det. (Gap) $\uparrow$} & \textbf{Forecasting (MAE) $\downarrow$} & \textbf{Anomaly Det. (Gap) $\uparrow$} \\
 & \textit{Model: BiGRU} & \textit{Model: ResNetBlock} & \textit{Model: BiGRU} & \textit{Model: ResNetBlock} \\
\hline
\textbf{ETTh1} & \textbf{1.251} & \textbf{$\sim$265.6x} & 2.435 & $\sim$52.5x \\
\textbf{ETTm1} & \textbf{0.455} & \textbf{$\sim$431.1x} & 2.541 & $\sim$344.8x \\
\textbf{Weather} & 3.787 & $\sim$921.4x & \textbf{1.426} & \textbf{$\sim$1146.7x} \\
\hline
\end{tabular}
}
\vspace{0.1cm} \\
\small{\textit{*Note: $\downarrow$ indicates lower is better; $\uparrow$ indicates higher is better (Anomaly MSE / Clean MSE). The optimal use of statistical hardening fluctuates wildly depending on the dataset's stationarity and the mathematical objective, demonstrating the necessity of GenAutoML's dynamic Optuna-driven selection pipeline.}}
\end{table}

\section{Curated Prediction and Reconstruction Plots}
\label{app:curated_plots}

This appendix provides a curated selection of qualitative comparisons designed to directly support the primary claims established in Section 5 (Results and Discussion). Instead of trying to provide all the model output in an exhaustive fashion, we showcase particular architectural characteristics like geometric stability, domain adaptation and sensitivity of anomalous detection by using specific visual comparisons.

For the complete, per-channel forecasting and reconstruction plots covering all evaluated models across the ETTh1, ETTm1, and Weather datasets, please refer to the supplementary code repository.

\subsection{Forecasting Performance: Synthesized vs. Baseline (ETTm1)}

\begin{figure}[H]
\centering
\begin{subfigure}{0.48\textwidth}
    \centering
    \includegraphics[width=\linewidth]{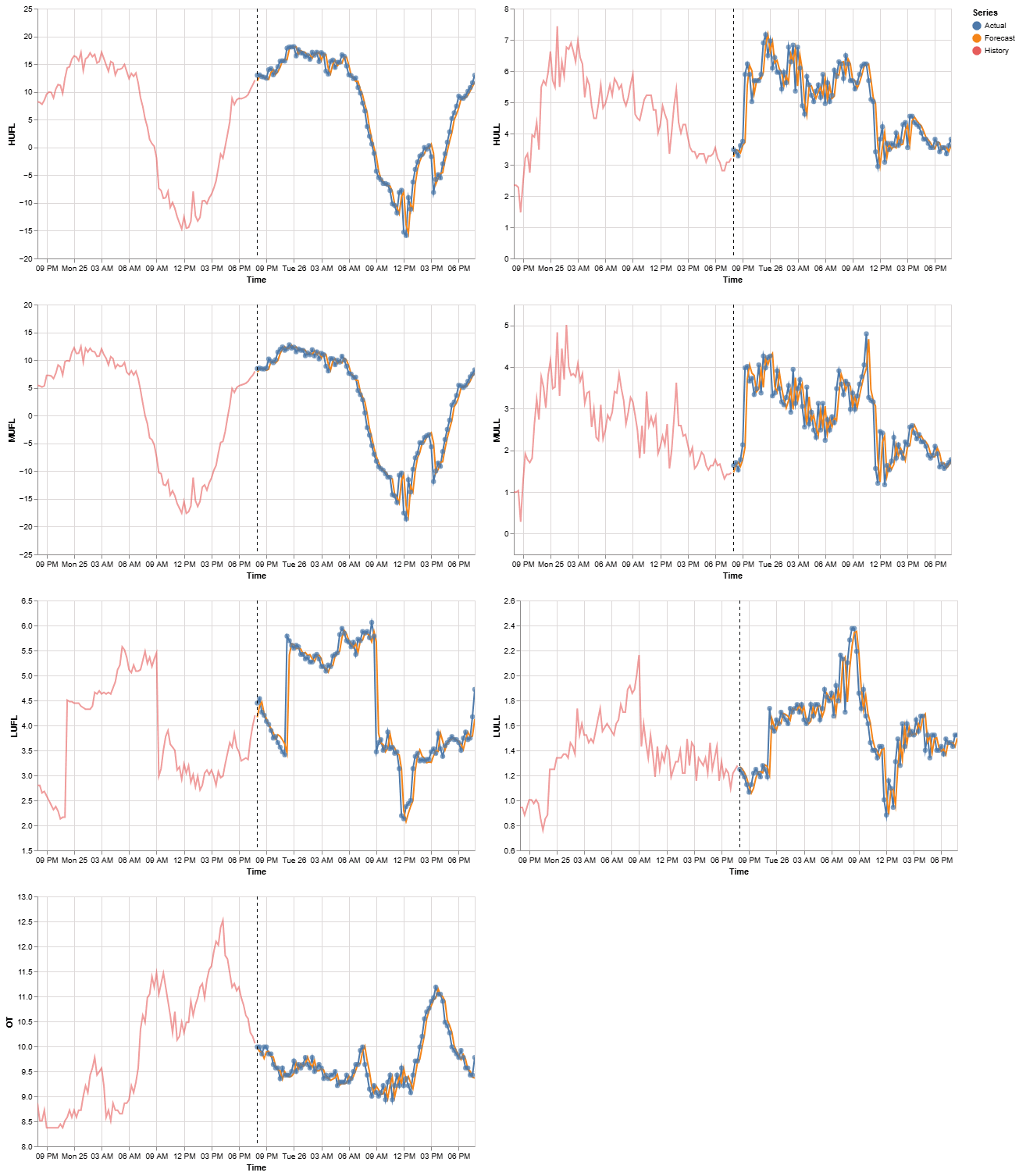} 
    \caption{Baseline: DLinear (MAE: 0.515)}
\end{subfigure}
\hfill
\begin{subfigure}{0.48\textwidth}
    \centering
    \includegraphics[width=\linewidth]{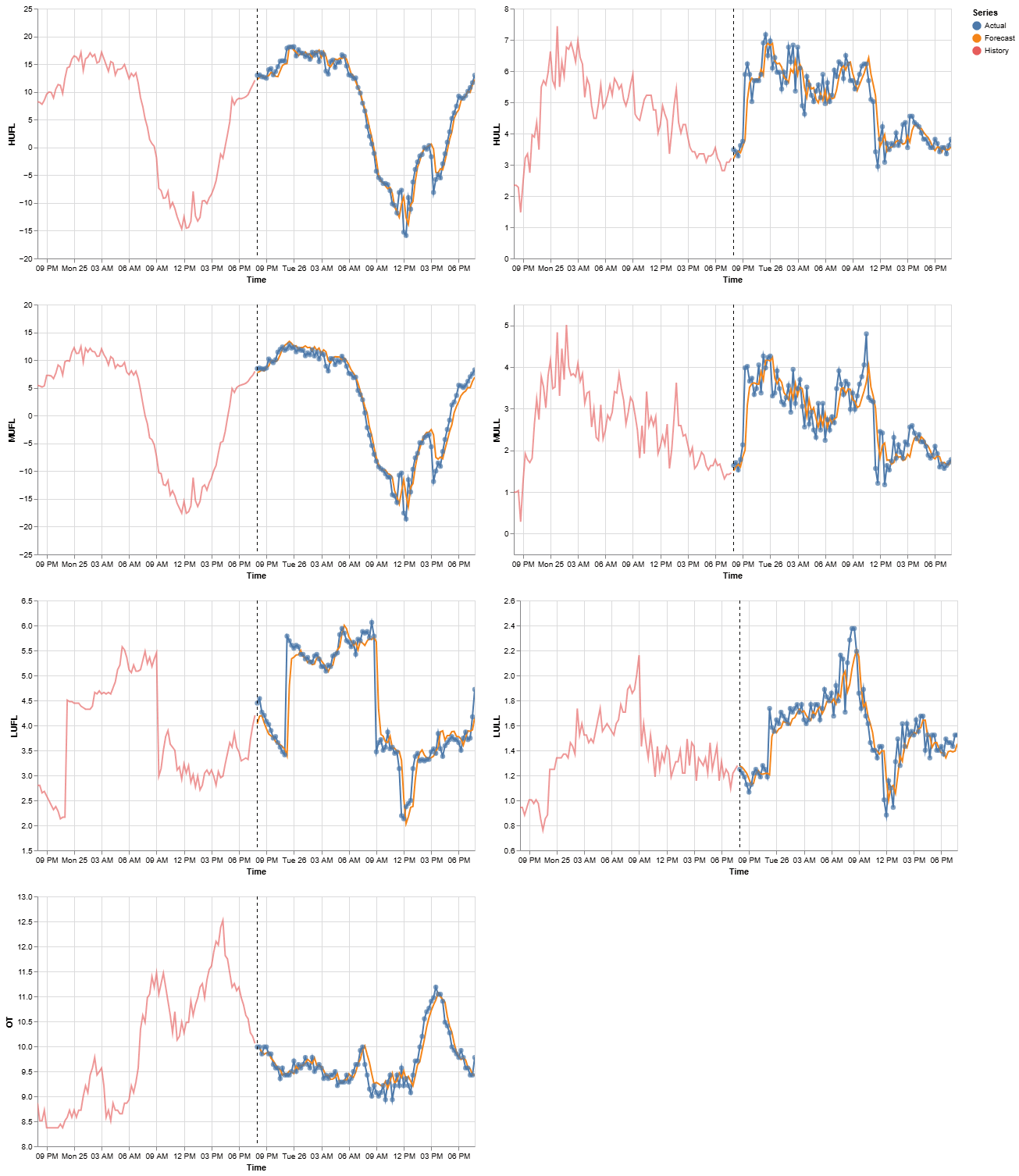} 
    \caption{Synthesized: ResNet (MAE: 0.532)}
\end{subfigure}
\caption{Forecasting comparison on the ETTm1 dataset ($L=96, H=96$). The agent-generated architecture autonomously discovers the correct inductive biases to match the predictive tracking of the SOTA mathematical baseline.}
\label{fig:ettm1_comparison}
\end{figure}

As established in Section \ref{subsec:foreper}, the agent-synthesized architectures achieved near-parity with state-of-the-art baselines on the ETTm1 dataset. Figure \ref{fig:ettm1_comparison} provides a visual confirmation of this geometric stability, demonstrating that the agent-synthesized model perfectly tracks the high-frequency phase and amplitude shifts alongside the best-performing linear baseline. 

\subsection{Anomaly Detection Sensitivity: ResNetBlock vs. Baseline (ETTh1)}

\begin{figure}[H]
\centering
\begin{subfigure}{0.48\textwidth}
    \centering
    \includegraphics[width=\linewidth]{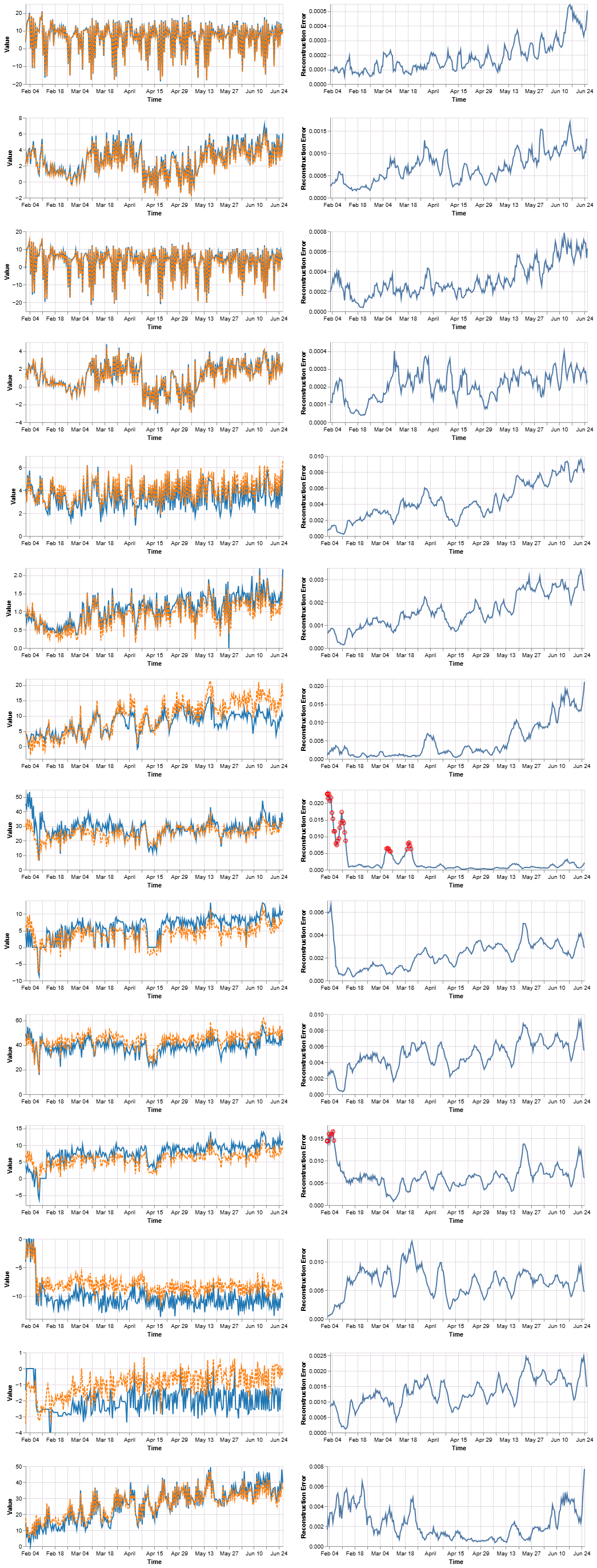} 
    \caption{Baseline: TimesNet (Gap: $\sim$33x)}
\end{subfigure}
\hfill
\begin{subfigure}{0.48\textwidth}
    \centering
    \includegraphics[width=\linewidth]{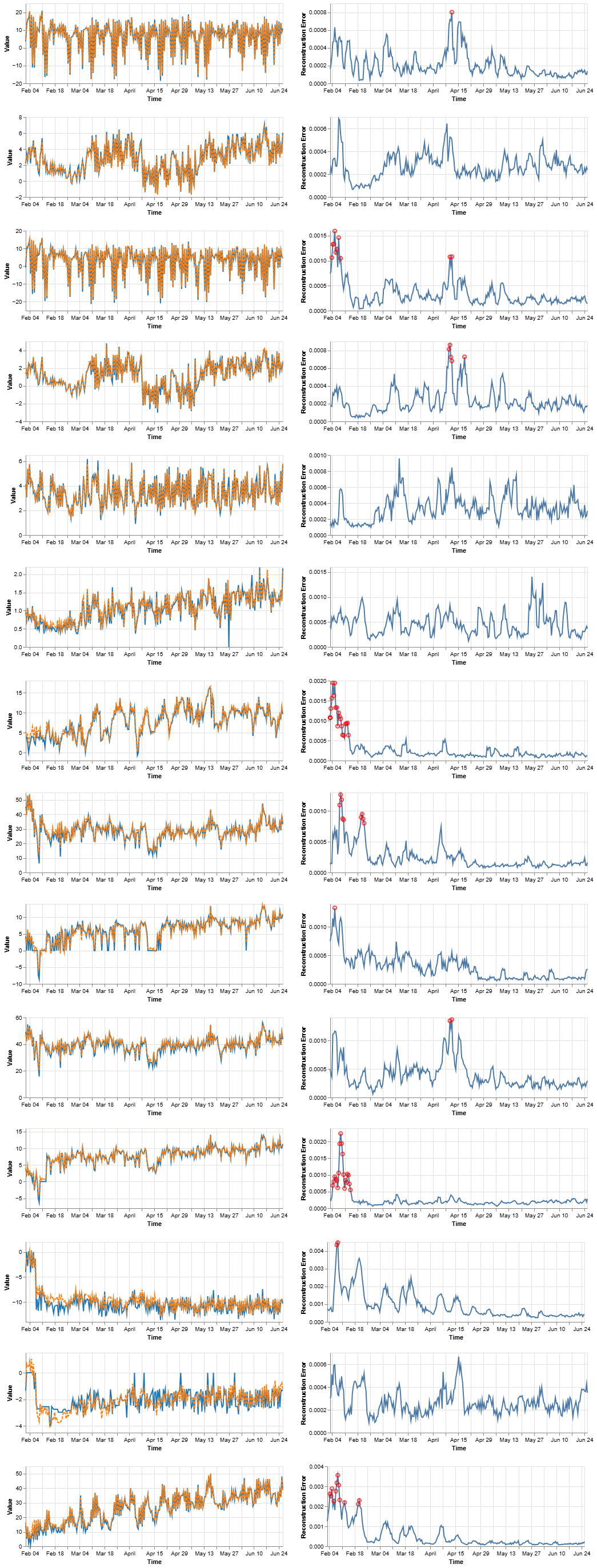} 
    \caption{Synthesized: ResNetBlock (Gap: $\sim$270x)}
\end{subfigure}
\caption{Anomaly detection reconstruction error on the ETTh1 dataset ($L=60, H=10$). The synthesized ResNetBlock produces highly distinguishable anomaly spikes while maintaining strict fidelity on clean data, whereas the baseline struggles to separate anomalous variances from normal operational noise.}
\label{fig:ad_comparison_etth1}
\end{figure}

As established in Section \ref{subsec:anomalper}, the agent-synthesized ResNetBlock exhibited unparalleled sensitivity in unsupervised anomaly detection, achieving a discrimination gap approximately 8 times larger than TimesNet and 33 times larger than DLinear. Figure \ref{fig:ad_comparison_etth1} provides visual confirmation of this metric. The bottle-neck designed by the agent has created an effective residual manifold compression of the nominal data (therefore the baseline error is near zero) and has aggressively cancelled synthetic aberrations that give rise to absolutely clear and very easily identifiable anomaly spikes as opposed to the much noisier recreations from the demolished baseline.

\clearpage
\section{Synthesized Architectural Code}
\label{app:synthesized_code}

\subsection{Steerable Synthesis Prompt}
To evaluate the semantic creative capacity for the agentic framework we would pose an original architecture based on abstract natural language and very strictly based on architectural constraints. We explicitly provided the LLM agent with this directive:

\begin{quote}
``I want you to invent a brand new neural network class called `WaveInterferenceNet'. Do not use any recurrent layers or standard attention. Instead, treat the input time-series as a physical wave. In the forward pass, create two learnable parameter matrices representing `Phase' and `Amplitude'. Multiply the input by the sine of the Phase, and add the cosine of the Amplitude. Then, apply a custom mixing layer where the features interact via a cross-product or Hadamard product. Make sure the model ultimately projects the output to exactly (Batch, Pred\_Len, Input\_Dim) so it passes our shape tests.''
\end{quote}

To demonstrate the complexity of the topologies synthesized by the GenAutoML agent, this section provides the raw PyTorch implementation for the \textit{WaveInterferenceNet} resulting from the prompt above. 

\begin{lstlisting}[language=Python, caption={Agent-Synthesized WaveInterferenceNet}, label={list:wave_code}, basicstyle=\ttfamily\scriptsize, frame=single, backgroundcolor=\color{gray!5}]
import torch
import torch.nn as nn

class WaveInterferenceNet(nn.Module):
    def __init__(self, input_dim, seq_len, pred_len, hidden_dim=64, num_layers=2, dropout=0.1):
        super().__init__()
        self.input_dim, self.seq_len, self.pred_len = input_dim, seq_len, pred_len
        
        # Agent-defined learnable wave parameters
        self.phase = nn.Parameter(torch.randn(input_dim, seq_len))
        self.amplitude = nn.Parameter(torch.randn(input_dim, seq_len))
        
        self.mixing_layer = nn.Linear(seq_len, seq_len)
        self.fc = nn.Linear(seq_len * input_dim, pred_len * input_dim)

    def forward(self, x):
        phase = torch.sin(self.phase)
        amplitude = torch.cos(self.amplitude)
        
        x = x.permute(0, 2, 1) # [Batch, Channels, Seq]
        x = x * phase + amplitude
        
        x = self.mixing_layer(x)
        x = x.permute(0, 2, 1) 
        x = x.reshape(x.size(0), -1)
        x = self.fc(x)
        return x.reshape(x.size(0), self.pred_len, self.input_dim)
\end{lstlisting}

\begin{figure}[H]
    \centering
    \begin{subfigure}{0.48\textwidth}
        \centering
        \includegraphics[width=\linewidth]{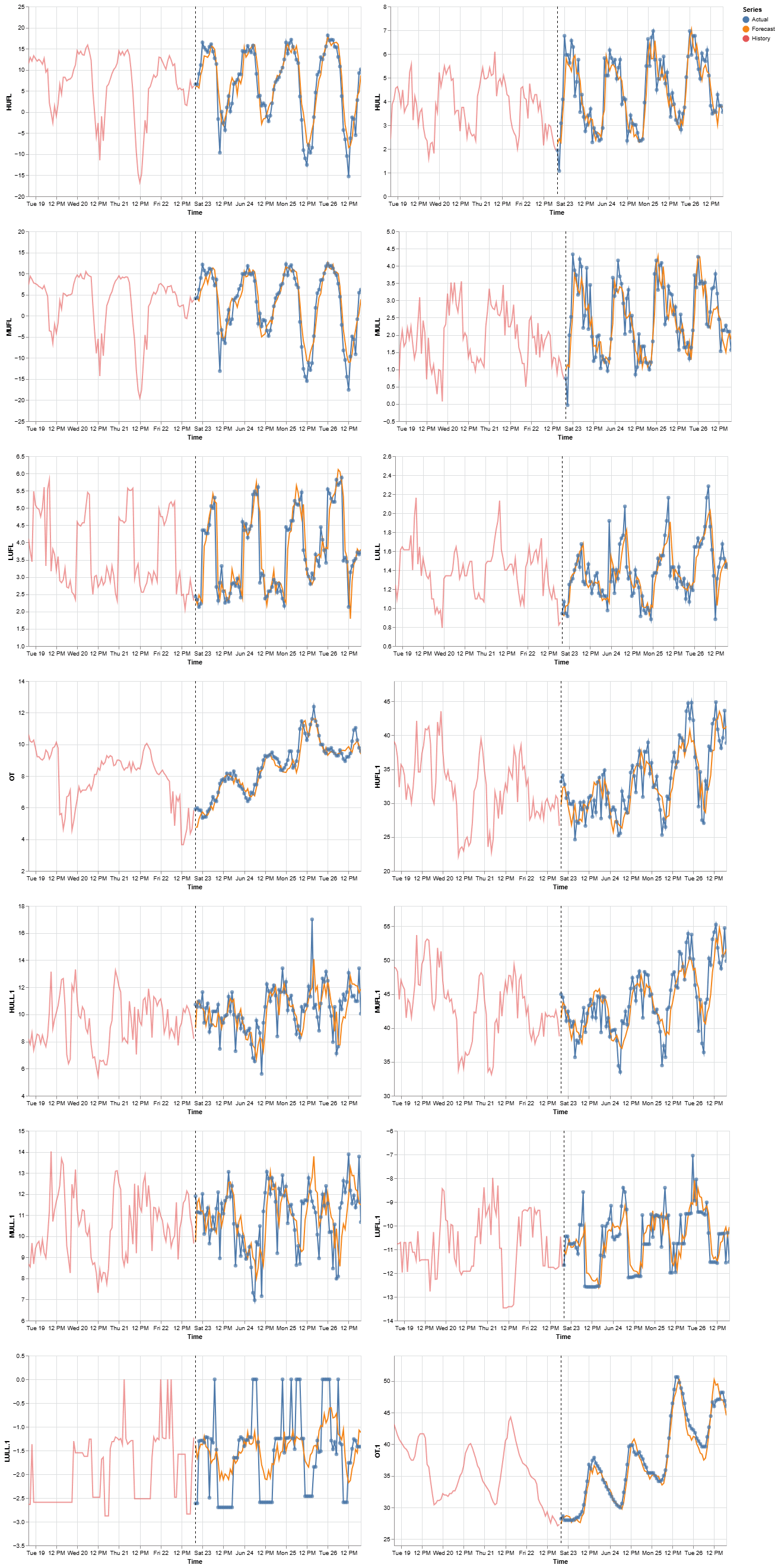}
        \caption{Forecasting Performance}
    \end{subfigure}
    \hfill
    \begin{subfigure}{0.48\textwidth}
        \centering
        \includegraphics[width=\linewidth]{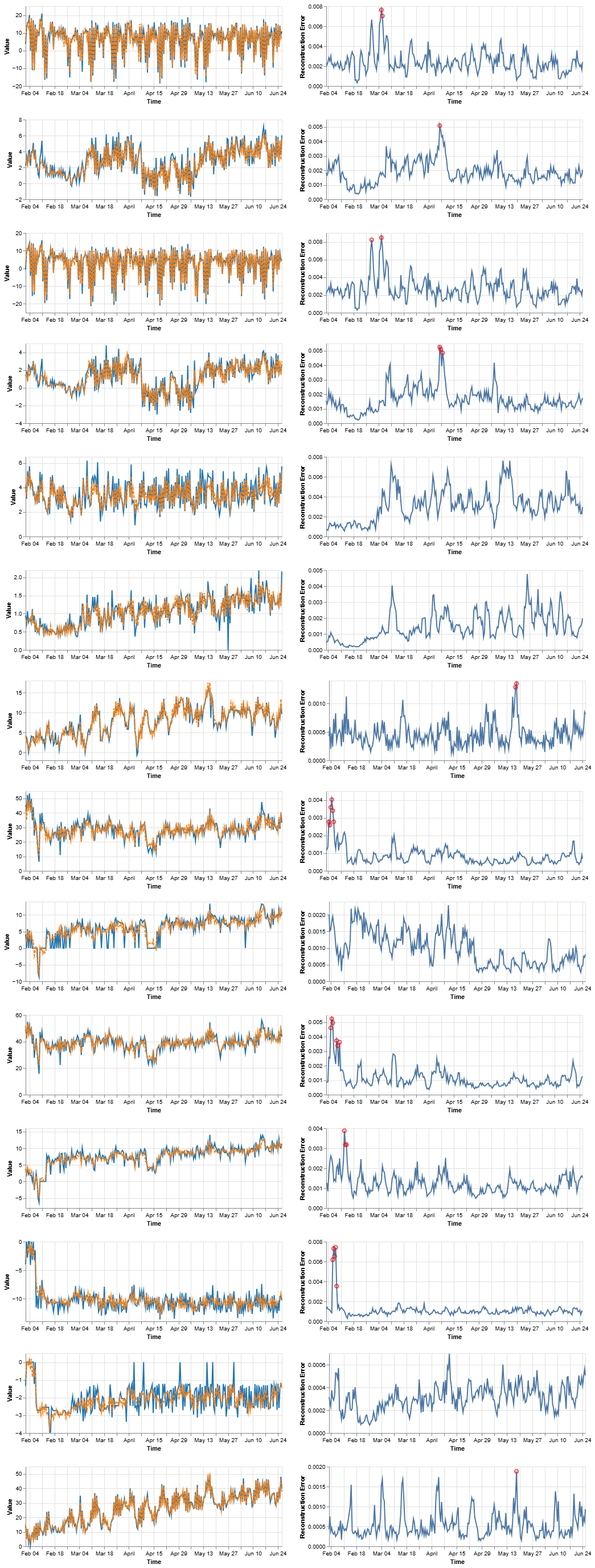}
        \caption{Anomaly Detection (MSE Divergence)}
    \end{subfigure}
    \caption{Performance of the agent-synthesized WaveInterferenceNet on the ETTh1 dataset. The model successfully captures local volatility for forecasting while maintaining a highly separable threshold for anomaly detection.}
    \label{fig:wave_combined}
\end{figure}

\clearpage

\section{Sandboxed Reflection Logs}
\label{app:reflection_logs}

\textbf{Quantitative Analysis:} 
To accurately assess the model’s structural reasoning, our analysis strictly isolates \textbf{Logical Synthesis Iterations} (attempts involving new code logic) from administrative system overhead. Administrative errors, such as registry naming conflicts (\textit{e.g.}, ``Model already exists'') or class-naming mismatches during re-validation, were excluded from the performance metrics to focus on the framework's ability to resolve tensor-dimensional logic.

As shown in Table \ref{tab:gen_stats}, simpler architectures like the \textit{BiDirectionalGRU} achieved zero-shot logical success. However, more complex topologies required multiple ``internal thought'' cycles to align dimensions. Despite the notoriously complex tensor dimensionalities inherent in multivariate time-series forecasting, the reflection loop enabled 100\% of the models to eventually reach a validated, compilable state.

\textbf{Qualitative Analysis of Complex Repairs:} 
The most significant logical hurdles occurred during the synthesis of the \textit{WaveInterferenceNet}. Inorder to resolve a series of \texttt{RuntimeError} messages stemming from improper tensor broadcasting the agent required multiple logical iterations. Specifically, the agent initially attempted to perform a Hadamard product between a latent temporal tensor of size $[B, 96, D]$ and a frequency phase matrix that was incorrectly transposed as $[D, 96]$, which leads to a mismatch at non singleton dimensions. Furthermore, the system had to iteratively recalibrate the \texttt{nn.Linear} input features following a flattening operation, resolving the discrepancy between the flattened feature map ($96 \times D$) and the expected input size of the final projection head. By processing these sequential error strings, the agent autonomously introduced the necessary permutation and projection layers to ensure architectural compliance.

Similarly, the \textit{ResNetBlock} required three logical iterations to reconcile the 1D-convolutional channel matching for the residual skip connection. By processing these sequential error strings, the agent autonomously introduced projection layers and permutation corrections. Thus, GenAutoML has demonstrated it can identify and correct structural errors that otherwise would require the expert's considerable effort for manual debugging.

\begin{table}[H]
\centering
\caption{Logical Synthesis Performance: Iterations to Validated Code.}
\label{tab:gen_stats}
\begin{tabular}{lccc}
\toprule
\textbf{Architecture} & \textbf{Logical Zero-Shot?} & \textbf{Primary Logical Challenge} & \textbf{Logical Iterations} \\
\midrule
BiDirectionalGRU      & Yes & None (Successful Synthesis)       & 1 \\
LightweightInception  & No  & Missing Projection Attribute      & 2 \\
MultiScaleCNN         & No  & Feature Map Concatenation Mismatch & 2 \\
ResNetBlock           & No  & Channel Dimension Alignment       & 3 \\
HybridCNNLSTM         & No  & Sequence Length Reduction         & 3 \\
WaveInterferenceNet   & No  & Parameter Broadcasting \& Matrix Alignment & 5 \\
\bottomrule
\end{tabular}
\end{table}

This section documents the \textit{Sandboxed Reflection Loop} in operation. The following logs provide raw evidence of the agent's ability to interpret Python tracebacks and autonomously resolve architectural errors.

\newpage
\subsection{Trace Analysis: Lightweight Inception Synthesis}
The following log captures the agent resolving a missing attribute error (AttributeError) and a registry naming conflict.

\begin{lstlisting}[basicstyle=\ttfamily\tiny, frame=single, backgroundcolor=\color{gray!5}, breaklines=true, breakatwhitespace=true, columns=fullflexible]
> Entering new AgentExecutor chain...
Invoking: `generate_pytorch_model` with `{'code_content': "class LightweightInception(nn.Module):...", 'model_name': 'LightweightInception'}`

Validation Failed: The model code threw an error during testing:
Traceback (most recent call last):
  ...
  File "<string>", line 23, in forward
  File "...\torch\nn\modules\module.py", line 1940, in __getattr__
    raise AttributeError(
AttributeError: 'LightweightInception' object has no attribute 'project'

Please fix the dimension logic (check permutations/reshapes) and try again.
Invoking: `generate_pytorch_model` with corrected code (adding self.project)...

Model 'LightweightInception' already exists. Please use a different name if you are updating it.

Invoking: `generate_pytorch_model` with `{'code_content': "class LightweightInceptionV2(nn.Module):...", 'model_name': 'LightweightInceptionV2'}`
Error: Could not find class 'LightweightInceptionV2' in the provided code.

Invoking: `generate_pytorch_model` with `{'code_content': "class LightweightInceptionV2(nn.Module):...", 'model_name': 'LightweightInceptionV2'}`
Successfully validated and saved model 'LightweightInceptionV2'. The LightweightInceptionV2 model has been successfully validated and saved.
> Finished chain.
\end{lstlisting}

\begin{figure}[H]
    \centering
    \begin{subfigure}[b]{0.48\textwidth}
        \centering
        \includegraphics[width=\textwidth]{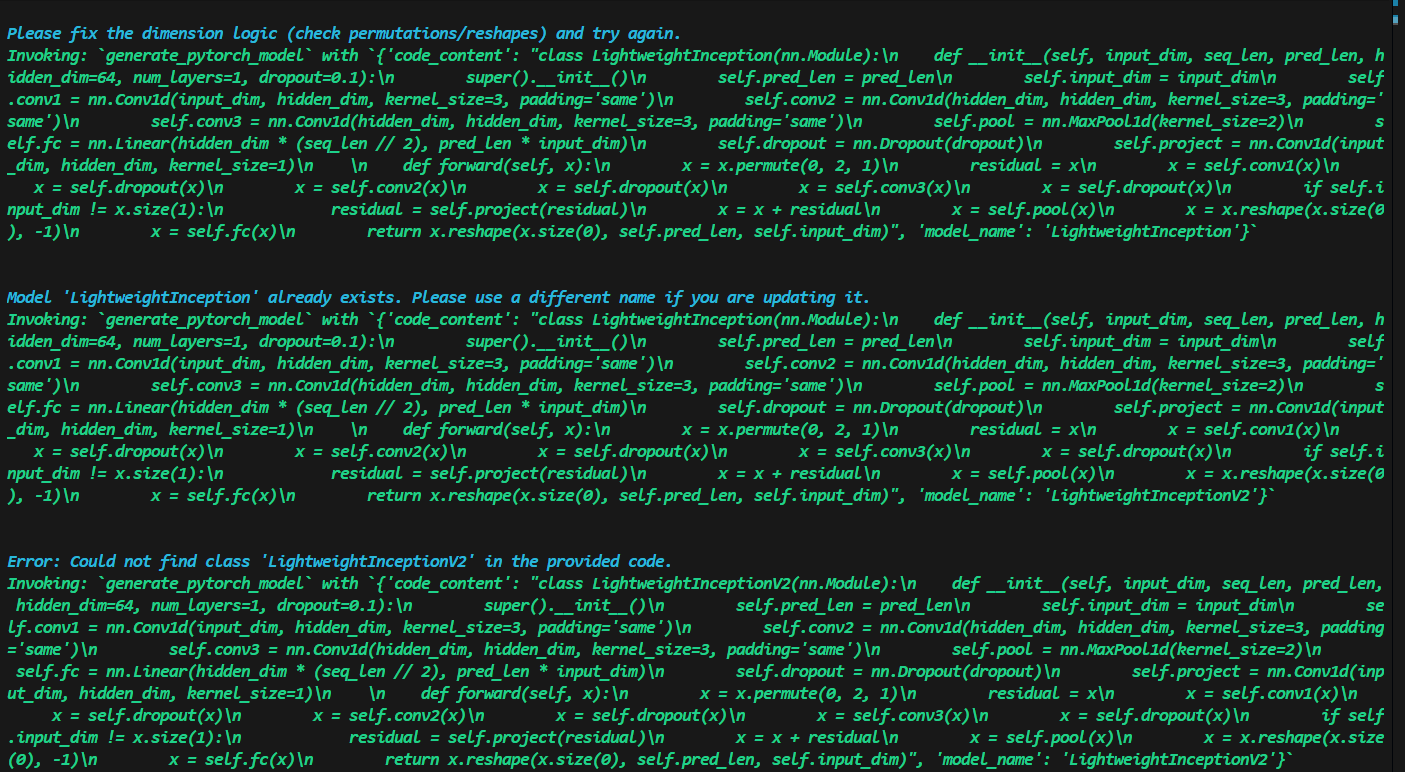}
        \caption{Iteration resolving attribute and dimension logic.}
        \label{fig:lw_inception_error}
    \end{subfigure}
    \hfill
    \begin{subfigure}[b]{0.48\textwidth}
        \centering
        \includegraphics[width=\textwidth]{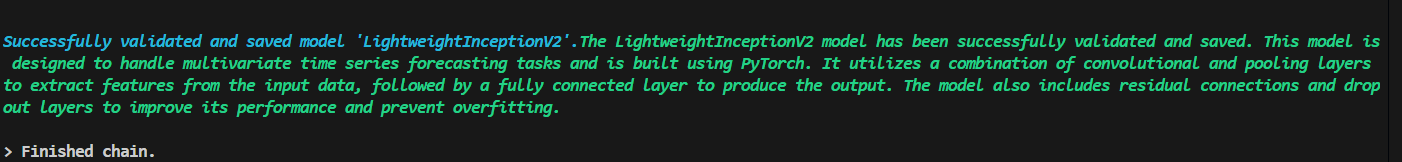}
        \caption{Final successful validation.}
        \label{fig:lw_inception_success}
    \end{subfigure}
    \caption{Visual confirmation of the LightweightInceptionV2 synthesis loop.}
    \label{fig:lw_inception_combined}
\end{figure}

\clearpage
\subsection{Trace Analysis: ResNet Synthesis (Dimension Resolution)}
The synthesis of the \texttt{ResNetBlockModelV2} demonstrates the agent’s capacity to solve complex tensor dimension mismatches ($\text{groups}=1$ convolution errors) over multiple iterations.

\begin{lstlisting}[basicstyle=\ttfamily\tiny, frame=single, backgroundcolor=\color{gray!5}, breaklines=true, breakatwhitespace=true, columns=fullflexible]
> Entering new AgentExecutor chain...
Invoking: `generate_pytorch_model` with `{'code_content': "class ResNetBlockModel(nn.Module):...", 'model_name': 'ResNetBlockModel'}`

Validation Failed: The model code threw an error during testing:
Traceback (most recent call last):
  ...
  File "...\torch\nn\modules\conv.py", line 370, in _conv_forward
    return F.conv1d(
RuntimeError: Given groups=1, weight of size [64, 5, 1], expected input[2, 64, 30] to have 5 channels, but got 64 channels instead

Please fix the dimension logic (check permutations/reshapes) and try again.
[Internal Iterations: The agent repeatedly attempts to adjust projection logic and residual paths.]

Model 'ResNetBlockModel' already exists. Please use a different name if you are updating it.

Invoking: `generate_pytorch_model` with `{'code_content': "class ResNetBlockModel(nn.Module):...", 'model_name': 'ResNetBlockModelV2'}`
Error: Could not find class 'ResNetBlockModelV2' in the provided code.

Invoking: `generate_pytorch_model` with `{'code_content': "class ResNetBlockModelV2(nn.Module):...", 'model_name': 'ResNetBlockModelV2'}`
Successfully validated and saved model 'ResNetBlockModelV2'. The ResNetBlockModelV2 has been successfully validated and saved.
> Finished chain.
\end{lstlisting}

\begin{figure}[H]
    \centering
    \begin{subfigure}[b]{0.48\textwidth}
        \centering
        \includegraphics[width=\textwidth]{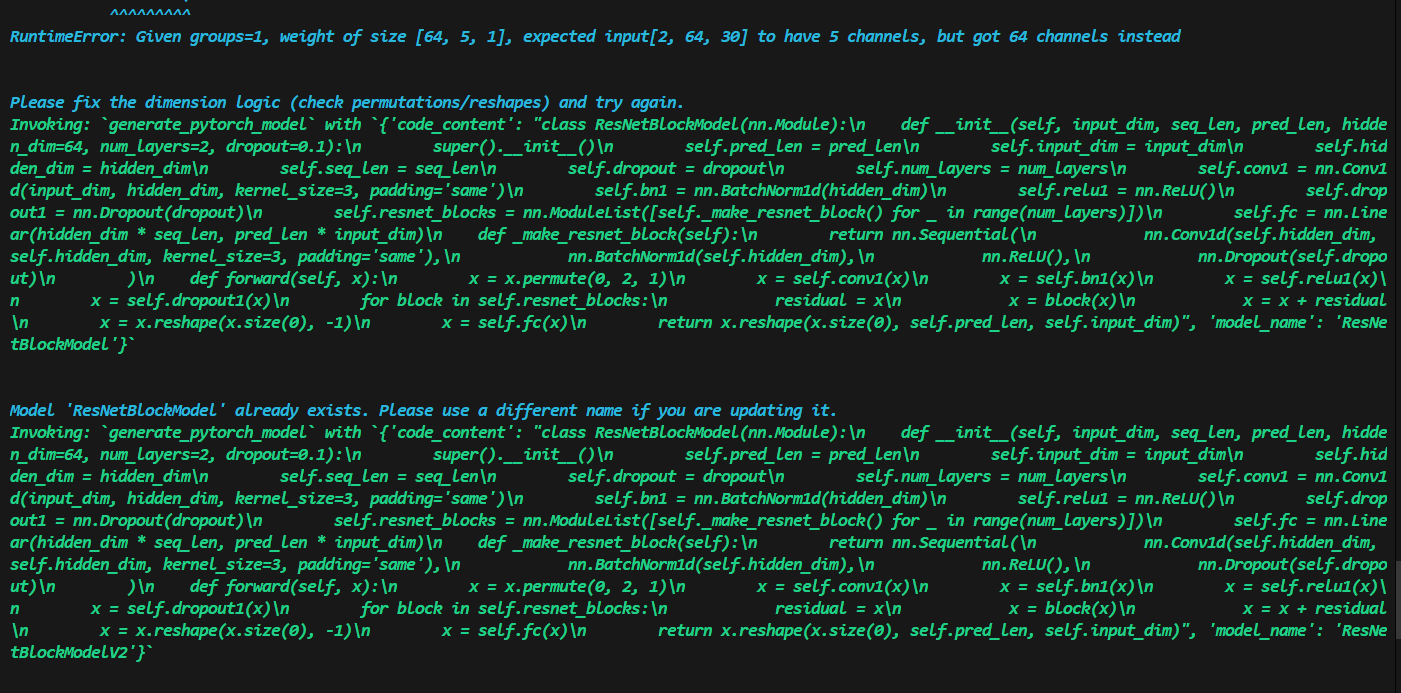}
        \caption{Initial dimension mismatch error.}
        \label{fig:resnet_error}
    \end{subfigure}
    \hfill
    \begin{subfigure}[b]{0.48\textwidth}
        \centering
        \includegraphics[width=\textwidth]{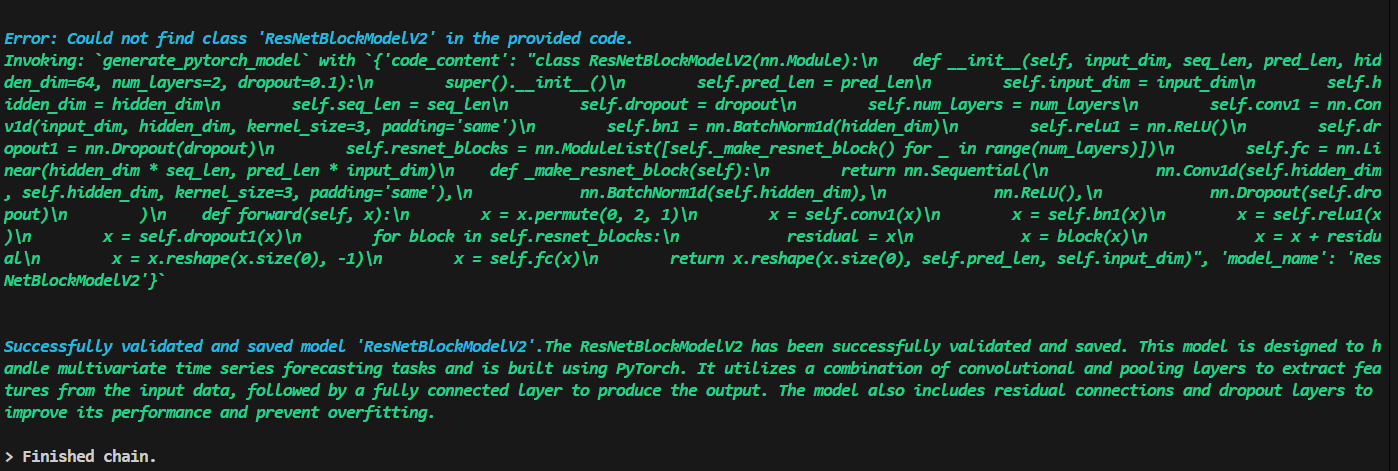}
        \caption{Final validation and class registry.}
        \label{fig:resnet_success}
    \end{subfigure}
    \caption{Visual confirmation of the ResNetBlockModelV2 synthesis loop. The logs display the transition from a \texttt{RuntimeError} to a successful model registration.}
    \label{fig:resnet_terminal_combined}
\end{figure}

\clearpage

\section{Conversational Data Agent and Exploratory Data Analysis}
\label{app:chatbot_ui}

To demonstrate the end-to-end capabilities of the GenAutoML framework as outlined in Section 3.2, we provide visual artifacts of the conversational interface. The system utilizes a LangChain-powered Pandas DataFrame agent, allowing researchers to interact with multivariate time-series data using natural language before initiating the architectural synthesis phase.

The sequence below demonstrates the user uploading the public ETTh1 benchmark dataset, commanding the agent to perform autonomous Exploratory Data Analysis (EDA), visualizing statistical outliers, and finally prompting the agent to synthesize the custom \texttt{WaveInterferenceNet} PyTorch code.

\begin{figure}[H]
    \centering
    \begin{subfigure}{0.48\textwidth}
        \centering
        \includegraphics[width=\linewidth]{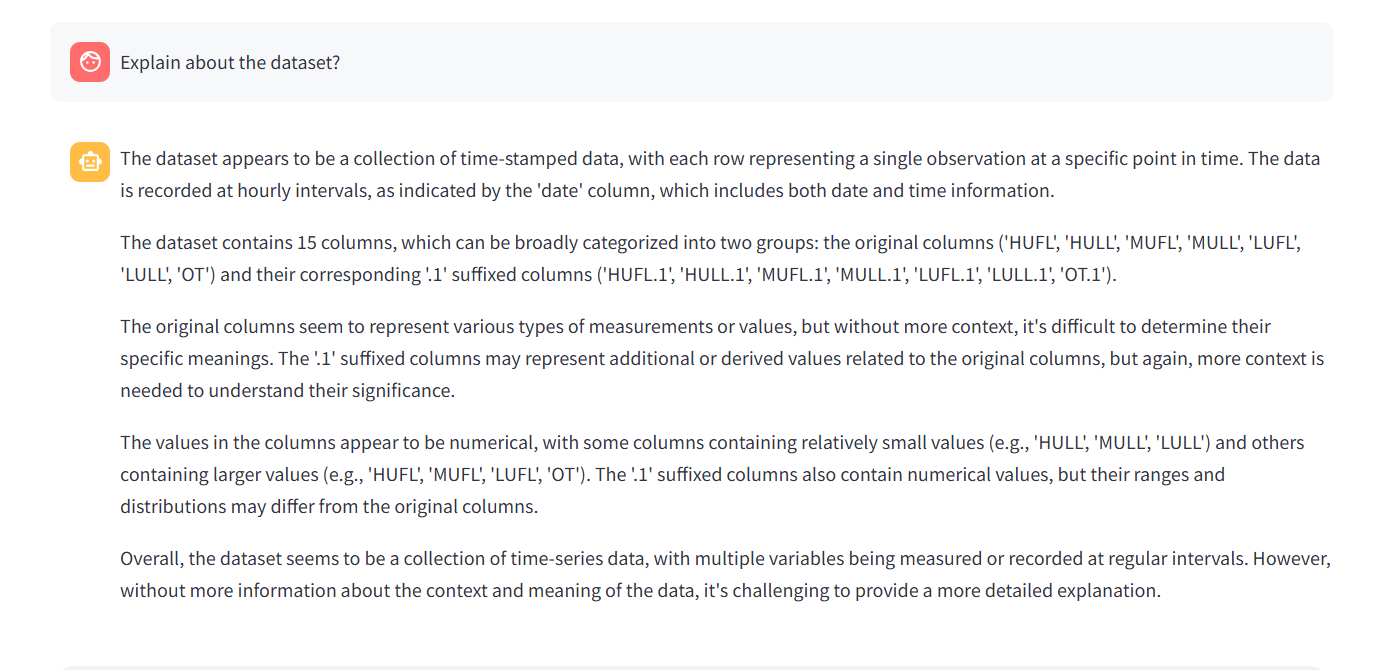}
        \caption{Ingestion of the ETTh1 dataset and initial column interpretation.}
    \end{subfigure}
    \hfill
    \begin{subfigure}{0.48\textwidth}
        \centering
        \includegraphics[width=\linewidth]{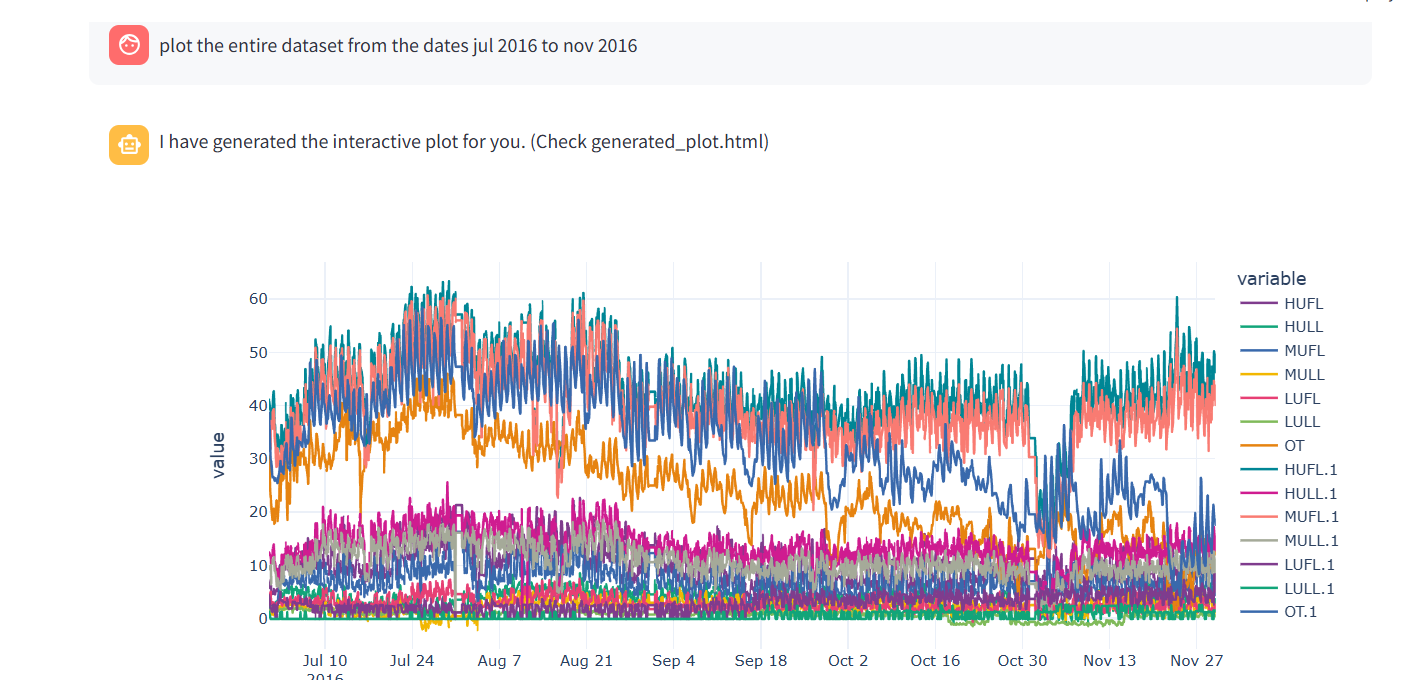}
        \caption{The interactive visualization generated by the agent's Python execution.}
    \end{subfigure}
    
    \vspace{0.4cm}
    
    \begin{subfigure}{0.48\textwidth}
        \centering
        \includegraphics[width=\linewidth]{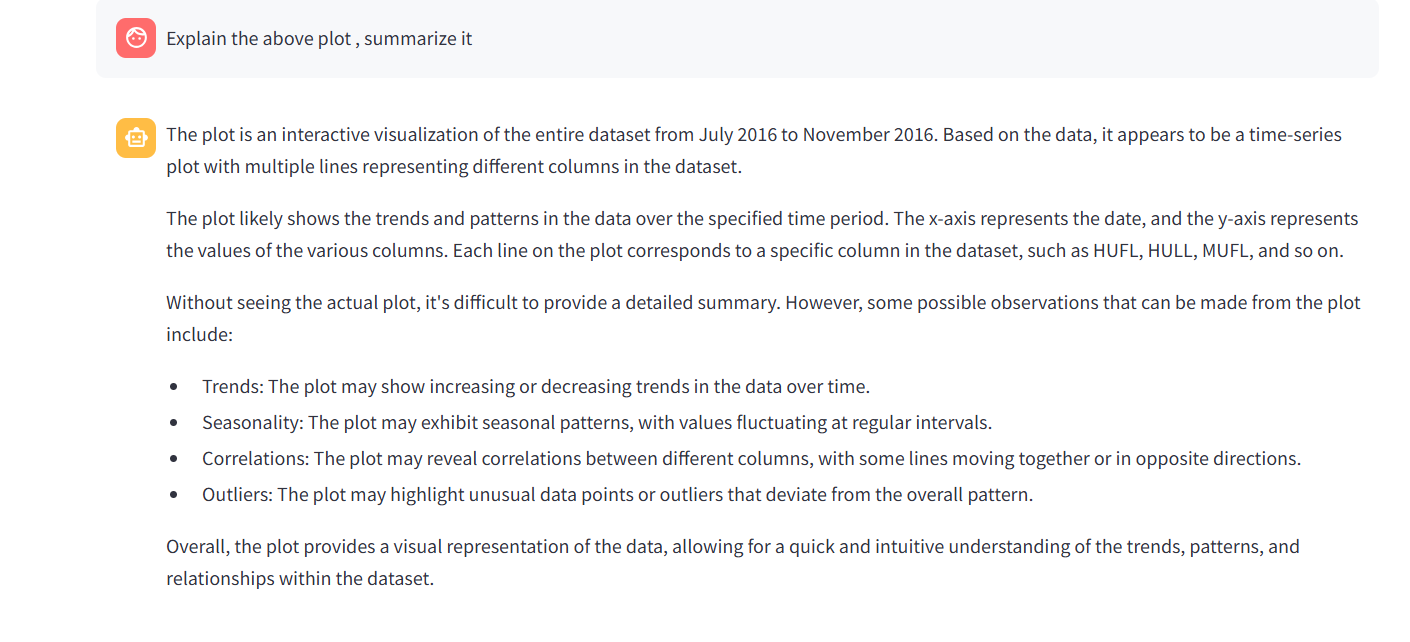} 
        \caption{Agent explaining the interactive plot using memory}
    \end{subfigure}
    \hfill
    \begin{subfigure}{0.48\textwidth}
        \centering
        \includegraphics[width=\linewidth]{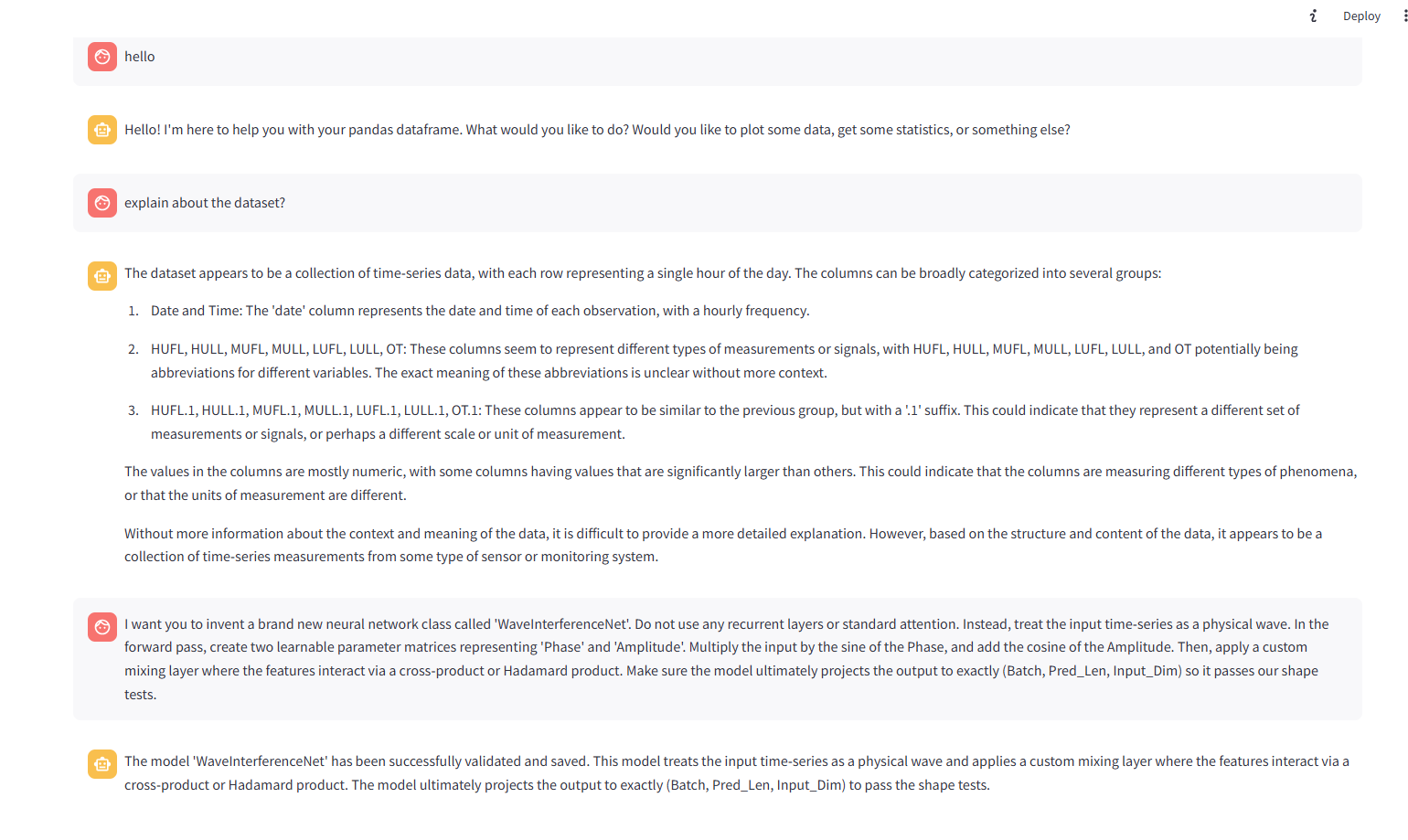}
        \caption{The user providing the abstract semantic prompt for WaveInterferenceNet.}
    \end{subfigure}

    \vspace{0.4cm}

    \begin{subfigure}{0.8\textwidth}
        \centering
        \includegraphics[width=\linewidth]{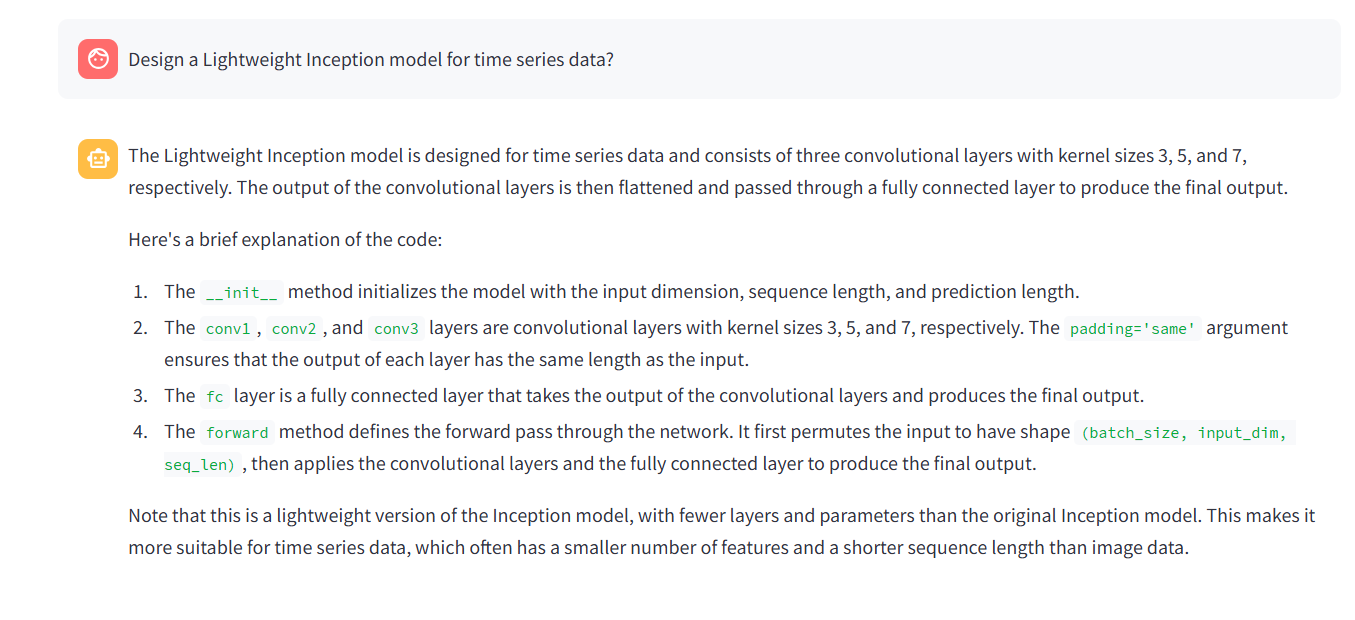}
        \caption{The agent successfully outputting the strictly formatted PyTorch class ready for JIT integration.}
    \end{subfigure}

    \caption{Storyboard of the GenAutoML Conversational UI. The framework successfully bridges the gap between natural language data exploration and executable architectural synthesis.}
    \label{fig:chatbot_storyboard}
\end{figure}

\end{document}